\documentclass[journal]{IEEEtran}
\usepackage{amsmath,amsfonts}
\usepackage{algorithmic}
\usepackage{algorithm}
\usepackage{array}
\usepackage[caption=false,font=normalsize,labelfont=sf,textfont=sf]{subfig}
\usepackage{textcomp}
\usepackage{stfloats}
\usepackage{url}
\usepackage{verbatim}
\usepackage{graphicx}
\usepackage{cite}
\usepackage[table]{xcolor} 
\usepackage{booktabs} 
\usepackage{multirow}
\usepackage{booktabs}
\usepackage{graphicx} 
\usepackage[colorlinks=true]{hyperref}

\usepackage{xcolor}
\definecolor{revisionred}{RGB}{200,0,0}

\hypersetup{
    colorlinks=true,
    citecolor=blue,     
    linkcolor=blue,     
}

\newcommand{\best}[1]{\textcolor{red}{#1}}
\newcommand{\second}[1]{\textcolor{blue}{#1}}

\definecolor{firstcolor}{HTML}{FDAE6B}
\definecolor{secondcolor}{HTML}{FDD0A2}
\definecolor{thirdcolor}{HTML}{FEE6CE}
\definecolor{restcolor}{HTML}{FFF5EB}

\begin{document}

\title{Gaussian Light Field Splatting: A Physical Prior-Driven Vision Transformer for Low-Light Image Enhancement}

\author{Yuhan~Chen, 
        Wenxuan~Yu, 
        Guofa~Li,~\IEEEmembership{Senior Member,~IEEE}, 
        Fuchen~Li,
        Kunyang~Huang, 
        Yicui~Shi, 
        Ying~Fang, 
        Wenbo~Chu, 
        and Keqiang~Li

\thanks{This work was supported by the National Natural Science Foundation of China under Grant 52672529 and 52372377, and the Young Beijing Scholars Program under Grant 2024-069. (\textit{Corresponding author: Guofa Li.})}%
\thanks{Yuhan~Chen, Wenxuan~Yu, Guofa~Li, Ying~Fang, and Yicui~Shi are with the College of Mechanical and Vehicle Engineering, Chongqing University, Chongqing 400044, China (e-mail: 20240701028@stu.cqu.edu.cn; wenxuanyu@cqu.edu.cn; liguofa@cqu.edu.cn; yingfang@stu.cqu.edu.cn; yicuishi@cqu.edu.cn).}%
\thanks{Fuchen Li is with the Herbert Wertheim College of Engineering, University of Florida, Gainesville, FL, USA (e-mail: fuchen.li@ufl.edu).}
\thanks{Kunyang Huang is with the Department of Electrical and Computer Engineering, Carnegie Mellon University, Moffett Field, CA 94035, USA (e-mail: kunyangh@andrew.cmu.edu).}%
\thanks{Wenbo~Chu is with the Western China Science City Innovation Center of Intelligent and Connected Vehicles (Chongqing) Co., Ltd., Chongqing 401329, China; the National Innovation Center of Intelligent and Connected Vehicles, Beijing 100176, China; the Engineering Research Center of Mechanical Testing Tech. and Equip. Ministry of Education, Chongqing University of Technology, Chongqing 400054, China (e-mail: chuwenbo@wicv.cn).}%
\thanks{Keqiang~Li is with the School of Vehicle and Mobility, Tsinghua University, Beijing 100084, China (e-mail: likq@tsinghua.edu.cn).}}

\markboth{IEEE TRANSACTIONS ON CIRCUITS AND SYSTEMS FOR VIDEO TECHNOLOGY}%
{Shell \MakeLowercase{\textit{et al.}}: A Sample Article Using IEEEtran.cls for IEEE Journals}


\maketitle

\begin{abstract}
Existing low-light image enhancement methods often suffer from local exposure imbalance and color distortion under complex, non-uniform illumination, while mainstream Vision Transformers lack explicit modeling of physics-based priors for illumination degradation. To address these limitations, we propose GLFS, a Gaussian Light Field Splatting-based Vision Transformer that incorporates continuous light-field priors from 2D Gaussian Splatting (2DGS) as an explicit physical bias into the Transformer self-attention mechanism to guide spatial gain estimation under complex non-uniform illumination. Specifically, GLFS represents scene illumination as a superposition of anisotropic Gaussian basis functions and introduces a physics-informed bias into the self-attention mechanism to adaptively infer a spatial gain field, enabling accurate and spatially uniform restoration under complex illumination conditions. To mitigate color shifts and structural degradation during enhancement, we further introduce a color vector angular loss and a luminance edge loss, with the former enforcing hue consistency and the latter improving the structural fidelity of local details. Extensive ablation studies and quantitative evaluations demonstrate that GLFS achieves competitive overall performance in illumination correction, structure preservation, and perceptual quality, validating the effectiveness of the continuous Gaussian light field prior for low-light image enhancement.
\end{abstract}


\section{Introduction}
\IEEEPARstart{L}{ow-light} image enhancement (LLIE) aims to recover clear image content from challenging visual conditions characterized by low illumination and high noise, serving as an essential preprocessing step for a wide range of high-level computer vision and pattern recognition tasks~\cite{ref1,ref2,ref3}. Deep learning-based LLIE methods have made substantial progress and have mainly evolved along two directions. One category relies on strictly aligned pairs of normal-light and low-light images for pixel-level supervision, typically achieving high reconstruction fidelity but being limited by the high cost of acquiring paired data and domain shift. The other category removes the dependence on paired data through adversarial learning, cycle consistency, or zero-reference physics-based losses, offering stronger generalization but lacking direct constraints on absolute exposure levels and color fidelity~\cite{ref10,ref21}. Both paradigms have inherent limitations. Therefore, combining paired supervision with physical constraints to maintain reconstruction accuracy while improving adaptability to complex scenes has become an important direction in LLIE.

Although existing LLIE methods, including Retinex-based unfolding networks~\cite{ref9,ref18}, implicit neural representations~\cite{ref24,ref25}, and diffusion models~\cite{ref22,ref23}, have improved overall image visibility to some extent, they still tend to suffer from local exposure imbalance under complex, spatially non-uniform illumination degradation. Architectures with strong long-range dependency modeling capabilities, such as state-space models~\cite{ref55} and Vision Transformers, have increasingly been adopted for image restoration. However, most of these architectures operate in a largely data-driven manner and lack explicit modeling of physics-based priors for illumination degradation. Without an explicit mechanism for modeling the physical characteristics of illumination degradation, these models may struggle to infer smooth spatial gain fields consistent with natural light field characteristics under extremely non-uniform illumination with severe spatial variations. This can lead to local artifacts and color distortions.

3D Gaussian Splatting (3DGS) has emerged as an effective technique for continuous scene and image representation, demonstrating strong capabilities in modeling geometry and appearance~\cite{ref28,ref30,ref51}. In particular, 2D Gaussian Splatting (2DGS) provides a high-fidelity mathematical formulation for representing continuous signals in the two-dimensional domain~\cite{ref49}. Several recent studies have introduced 2DGS into low-light image enhancement, using Gaussian basis functions to continuously model the diffusion and attenuation of complex illumination~\cite{ref47,ref48}. However, existing methods have not systematically investigated how to embed the continuous 2DGS representation as an explicit physical bias into the attention mechanism of Transformer architectures. Integrating the continuous physics-based illumination modeling mechanism of Gaussian splatting into Transformers therefore remains largely unexplored, particularly as an explicit physical prior for low-light enhancement.

Specifically, bridging 2DGS with self-attention involves three major technical challenges: (1) \emph{representation gap}: how to map continuous analytical Gaussian parameters onto discrete semantic tokens without sacrificing their physical meaning; (2) \emph{optimization conflict}: how to prevent rigid geometric constraints from interfering with the flexible semantic clustering of Vision Transformers, particularly during the early stages of training; and (3) \emph{numerical instability}: how to consistently maintain the positive semidefiniteness of Gaussian covariance matrices during frequent non-local feature updates.

To this end, we propose Gaussian Light Field Splatting (GLFS), a Vision Transformer (ViT) that integrates the light field prior of 2DGS into the Transformer self-attention mechanism, combining physics-based illumination constraints with the non-local modeling capability of Transformers. Specifically, instead of performing unconstrained feature mapping in the latent space as in conventional ViTs, GLFS employs a multi-scale Gaussian tokenizer to represent scene illumination as a superposition of anisotropic Gaussian basis functions parameterized by their means, covariances, and opacities. A physics-informed bias derived from Gaussian affinity is then explicitly introduced into the self-attention computation to guide the network in adaptively inferring a continuous and smooth spatial gain field, enabling spatially uniform restoration under complex, non-uniform illumination.

In addition, to address the color shifts and structural degradation commonly encountered during low-light enhancement, we further introduce two physical constraints in the color and gradient domains. Specifically, we propose a color vector angular loss that constrains the cosine similarity between the RGB vectors of corresponding pixels before and after enhancement. This formulation allows substantial changes in RGB vector magnitude associated with brightness while preserving the vector direction associated with hue, thereby reducing color distortions induced by the conventional gray-world assumption. Meanwhile, a luminance edge loss based on the Sobel operator is designed to preserve high-frequency gradients in the luminance Y channel while reducing interference from chrominance information, improving the structural fidelity of local details. The main contributions of this work are summarized as follows:

\begin{enumerate}
    \item We propose GLFS, a Vision Transformer based on Gaussian Light Field Splatting. To the best of our knowledge, GLFS is the first to explicitly integrate the light field prior of 2DGS into the Transformer self-attention mechanism, providing a physics-prior-driven representation for low-light image enhancement.
    
    \item We design a light field construction mechanism based on the superposition of Gaussian basis functions. By explicitly introducing anisotropic Gaussian biases into the self-attention computation, the proposed mechanism enables smooth and accurate inference of the spatial gain field under complex, non-uniform illumination, effectively mitigating local exposure imbalance.
    
    \item To improve the preservation of color consistency and structural information, we introduce a color vector angular loss and a luminance edge loss. As reference-free physical regularizers, these two losses further constrain the enhancement process in the color and gradient domains beyond paired reconstruction supervision. They improve hue consistency and local structural fidelity while providing additional gradients in regions where reference signals are unavailable or imperfectly aligned.
\end{enumerate}

\section{Related Work}

This section reviews two research areas closely related to our work. We first summarize the development of low-light image enhancement (LLIE), from traditional physics-based priors to deep neural networks. We then discuss advances in Gaussian splatting for explicit image representation, providing background for the physics-based design of GLFS.

\subsection{Low-Light Image Enhancement}

Low-light image enhancement (LLIE) aims to recover clear image content and faithful colors from images degraded by low illumination and noise, and serves as an important preprocessing step for many high-level computer vision tasks~\cite{ref1,ref2}. As large-scale benchmarks and evaluation protocols have matured~\cite{ref3}, LLIE research has evolved from traditional physics-based priors toward data-driven deep learning methods.

Early methods mainly relied on handcrafted physical constraints for illumination estimation. For example, LIME~\cite{ref5} estimates the overall illumination distribution of a scene within a conventional optimization framework by imposing a structural smoothness prior on an initial illumination map, without using deep feature representations. However, such methods are prone to local noise amplification and color distortion under extremely non-uniform illumination~\cite{ref57}. As deep learning advanced, supervised methods based on paired normal-light and low-light images became a major research direction. Retinex-Net~\cite{ref26} combines classical Retinex decomposition with data-driven convolutional neural networks (CNNs) to jointly learn illumination and reflectance components in an end-to-end manner, helping establish the supervised deep learning framework for low-light enhancement. Subsequently, a variety of CNN architectures were developed to improve feature extraction and reconstruction efficiency in complex scenes. For instance, FMR-Net~\cite{ref11} and FRR-NET~\cite{ref12} employ multi-scale residual feature aggregation and structural re-parameterization, respectively, reducing inference latency on mobile devices while maintaining reconstruction quality. EFINet~\cite{ref15} introduces an iterative feedback mechanism between enhancement and fusion to progressively refine high-frequency textures and structural information through hierarchical feature propagation. In addition, to address domain shift and limited generalization associated with training on synthetic paired datasets, R2RNet~\cite{ref27} learns a mapping from real low-light images to real normal-light images, while related paired instance learning methods~\cite{ref16} explore more robust degradation-to-restoration mappings from challenging real-world low-quality image pairs.

Alongside supervised approaches, unsupervised and zero-shot learning have also received considerable attention. EnlightenGAN~\cite{ref10} introduced an adversarial learning framework based on unpaired data. Retinex-based deep unfolding networks, such as Uretinex-Net~\cite{ref18}, which incorporates physical equations, and unfolding networks with collaborative prior architecture search~\cite{ref9}, convert conventional optimization procedures into learnable neural network layers, improving model interpretability and detail preservation under extremely dark conditions. RetinexDIP~\cite{ref69} combines the Retinex decomposition model with a deep image prior to jointly estimate illumination and reflectance components without external training data. Methods based on adaptive total variation~\cite{ref13} and zero-reference deep curve estimation, including ZeroDCE~\cite{ref6,ref7}, its accelerated variant~\cite{ref8}, and the optimal curve estimation method ChebyLighter~\cite{ref14}, perform flexible and efficient illumination correction using reference-free physics-based losses. Latent diffusion models~\cite{ref21,ref22,ref23} and implicit neural representations~\cite{ref24,ref25} have further improved the visual quality of unpaired enhancement by leveraging generative priors. To address the coupling between luminance attenuation and noise amplification under low-light conditions, Wang \textit{et al.}~\cite{ref71} extract separate luminance and noise priors to guide brightness restoration and noise suppression. Autoregressive denoising methods for real noise distributions~\cite{ref17,ref19} and compact lightweight architectures~\cite{ref4,ref20} have also improved the practical deployability of LLIE methods.

Purely supervised and purely unsupervised learning are not necessarily mutually exclusive. Purely supervised models are limited by the scene coverage of paired training data, whereas purely unsupervised objectives, such as the gray-world assumption and fixed exposure targets, lack absolute brightness and color references and may produce systematic shifts in scenes with strong color casts. Accordingly, some studies combine limited paired supervision with unpaired or reference-free constraints within a unified framework to improve reconstruction accuracy and distributional robustness. GLFS follows this hybrid supervision setting. It retains a bidirectional adversarial framework and cycle-consistency constraints, applies direct pixel-level and perceptual supervision to aligned image pairs, and uses physical regularization based on the Gaussian light field to constrain the continuity of intermediate physical quantities.

Transformer architectures have been increasingly applied to low-light image enhancement and general image restoration, with different studies modifying self-attention in terms of modeling domain and computational complexity. Jiang \textit{et al.}~\cite{ref70} propose a stage-wise Transformer-guided network that progressively models low-light degradation features across different scales and directions, enabling effective enhancement under complex illumination conditions. CSTG-CWHFlow~\cite{ref59} constructs Compressive Self-Attention by compressing features along the height and channel dimensions and incorporates Zero-Map Attention to balance global dependency modeling and zero-valued pixel recovery. LLFormer~\cite{ref60}, designed for ultra-high-resolution low-light image enhancement, employs Axis-Based Multi-Head Self-Attention to model dependencies along the two spatial axes, reducing the attention cost for high-resolution images. IIAG-CoFlow~\cite{ref61} jointly employs Inter-Channel Attention and Intra-Channel Attention to model inter-channel and intra-channel feature dependencies, respectively, and further incorporates Zero-Map Attention. In addition, Restormer~\cite{ref62}, developed for general high-resolution image restoration, introduces Multi-Dconv Head Transposed Attention, which computes self-attention along the channel dimension to efficiently capture long-range dependencies. These studies show that self-attention can support long-range interactions through spatial-axis decomposition, feature compression, and channel-domain modeling, with different trade-offs between modeling capability and computational complexity. In contrast to these data-driven attention designs, our focus is to use 2DGS to introduce explicit illumination geometry into the attention computation.

Despite the improved ability of the aforementioned Transformer architectures and the state-space model MambaLLIE~\cite{ref55} to model long-range dependencies, their feature interactions are still primarily governed by data-driven correlations and do not explicitly capture the continuous spatial diffusion and anisotropic attenuation of natural light fields. Under extremely non-uniform illumination with severe spatial variations, the absence of such physical illumination geometry may lead to unstable local gain estimation and exposure imbalance. Therefore, we further investigate how to embed the continuous Gaussian light field represented by 2DGS into self-attention as an explicit physical prior. This introduces physical constraints on spatial illumination distributions while preserving the non-local modeling capability of Transformers.

\subsection{Gaussian Splatting and Explicit Image Representation}

3D Gaussian Splatting (3DGS) has emerged as a continuous explicit scene representation technique, offering an alternative to conventional implicit radiance-field rendering approaches such as NeRF~\cite{ref29} and implicit coordinate networks with periodic activation functions~\cite{ref31}. Specifically, 3DGS~\cite{ref28} explicitly parameterizes 3D geometry and appearance using a set of anisotropic Gaussian primitives characterized by means, covariances, colors, and opacities. This analytical representation avoids the high computational overhead associated with ray marching and dense sampling in multilayer perceptron (MLP)-based implicit neural representations, while supporting real-time rendering with high geometric fidelity. To further improve the geometric quality and density of Gaussian reconstruction, various extensions have been developed. For example, GaussianPro~\cite{ref39} introduces a progressive propagation strategy guided by normal and depth information to guide the densification of Gaussian ellipsoids in complex geometric structures and textureless regions, reducing artifacts and holes in novel-view synthesis. 3DGS has subsequently been extended to a broad range of tasks, including high-fidelity modeling of dynamic urban street scenes~\cite{ref32}, efficient construction of 4D world models and scenes for autonomous driving~\cite{ref38,ref42,ref43}, real-time self-distilled rendering of large-scale urban scenes~\cite{ref40,ref41}, and cross-modal 3D content generation~\cite{ref44}. Moreover, continuous Gaussian representations have shown strong capabilities in modeling high-dimensional dynamic signals, including generalizable multi-view stereo reconstruction~\cite{ref33,ref34} and long-range spatiotemporal modeling of physical motion trajectories~\cite{ref35,ref36,ref37}.

As 3D Gaussian representations have advanced, anisotropic Gaussian basis functions have also demonstrated strong representational capacity for modeling continuous signals in two dimensions, leading to the development of 2D Gaussian Splatting (2DGS)~\cite{ref30}. In particular, for explicit image representation and efficient compression, GaussianImage~\cite{ref51} and its adaptive variant, Instant GaussianImage~\cite{ref50}, demonstrate that 2DGS can serve as a compact and continuous approach to general image parameterization. Unlike conventional discrete pixel grids, 2DGS achieves high-fidelity image reconstruction by directly optimizing the attributes of Gaussian primitives through backpropagation, providing a continuous analytical representation for local geometric variations, texture transitions, and color gradients. LIG~\cite{ref49} further introduces a hierarchical-resolution structure for 2DGS, alleviating the GPU memory and computational bottlenecks associated with explicit representation of ultra-high-resolution images. With its continuous and differentiable representation, 2DGS has been extended to various low-level vision and cross-modal alignment tasks, including sparse Gaussian-based dataset distillation~\cite{ref52}, compact representation of aligned features in vision-language models~\cite{ref53}, and high-fidelity image super-resolution at arbitrary scales~\cite{ref54}.

\begin{figure*}[t]
    \centering
    \includegraphics[width=\textwidth]{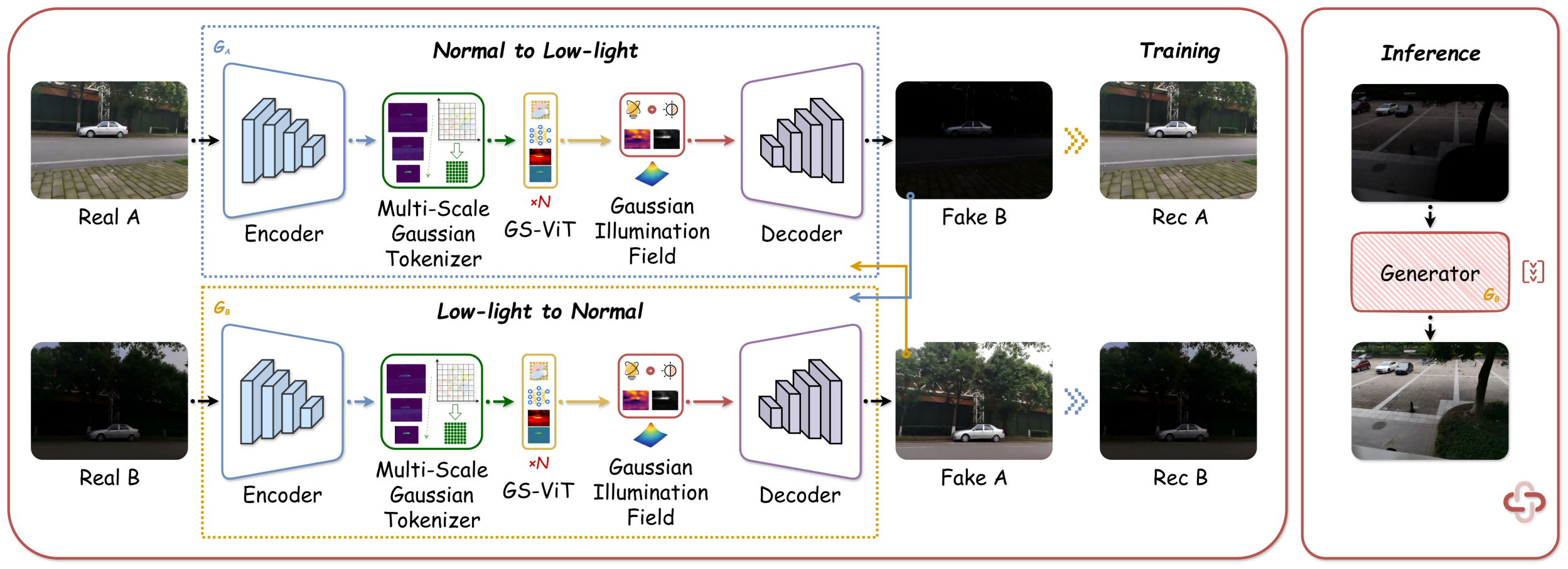}
    \caption{The overall training and inference framework of the proposed GLFS. GLFS adopts an encoder--decoder architecture with GS-ViT as the bottleneck, where explicit physical priors are embedded to guide the enhancement process.}
    \label{fig:overall_framework}
\end{figure*}

Beyond image-level representation, Gaussian splatting has also been explored for modeling more abstract sparse perceptual signals. For example, KGS-GCN~\cite{ref56} models discrete skeletal joints as anisotropic Gaussian distributions driven by instantaneous kinematic information and further constructs a probabilistic topology using statistical distances between joint Gaussian distributions. This formulation alleviates the representation bottlenecks caused by sparse sampling and fixed physical topology in action recognition. These developments indicate that Gaussian splatting has expanded beyond geometric rendering toward task-adaptive continuous prior modeling.

At the intersection of illumination estimation, physics-based degradation modeling, and low-light image processing, the continuous spatial characteristics and differentiability of Gaussian representations have also attracted increasing attention. LL-Gaussian~\cite{ref45} and LLGS~\cite{ref46} introduced Gaussian splatting constraints into sparse 3D reconstruction and enhancement in extremely dark environments, demonstrating the robustness of Gaussian representations in preserving spatial structure under severe illumination degradation and high noise levels. Recent studies have also explored the direct application of 2DGS to unsupervised low-light image enhancement. For example, LL-GaussianImage~\cite{ref47} systematically explores an efficient zero-shot enhancement framework based on 2DGS image reconstruction, while LL-GaussianMap~\cite{ref48} leverages the anisotropic attenuation of 2DGS to explicitly model continuous spatial illumination gain maps in natural images.

However, although these studies suggest a close correspondence between the spatial attenuation of Gaussian basis functions and the non-uniform distribution of natural light fields, they generally treat Gaussian representations as standalone fitting tools or isolated gain generators rather than integrating them into the underlying network architecture. In contrast, GLFS embeds the anisotropic light field prior of 2DGS as an explicit physics-informed bias into the self-attention computation of a Vision Transformer. This design preserves the non-local modeling capability of Transformers while introducing physical constraints for modeling complex spatial illumination degradation, enabling spatially uniform low-light restoration under hybrid supervision.

\section{Proposed Method}
\label{sec:proposed_method}

Low-light image enhancement is inherently a highly ill-posed inverse problem. The task aims to recover images with natural exposure, faithful colors, and clear structural details from observations affected by low illumination, noise, and spatially non-uniform lighting. In particular, under complex, non-uniform illumination, uncertainty in spatial gain estimation can lead to local exposure imbalance, hue shifts, and degradation of high-frequency details. As discussed in the Introduction, mainstream Vision Transformers (ViTs) provide strong long-range modeling capabilities, but their attention priors are typically based on relative positional information or learnable biases and are not explicitly tied to the physical formation process of scene illumination. In contrast, Retinex-based unfolding methods provide physical interpretability, but their linear decomposition formulation limits their ability to characterize complex anisotropic light fields. To overcome these limitations, we propose Gaussian Light Field Splatting (GLFS), which formulates low-light enhancement as a physically grounded reconstruction process explicitly driven by a continuous Gaussian light field and optimized under a hybrid supervision framework. In Section~\ref{sec:overall_framework}, we first introduce the overall architecture and mathematical formulation of GLFS. We then describe its three key physical-prior modules: the Multi-Scale Gaussian Tokenizer (MGT) in Section~\ref{sec:mgt}, GS-ViT with Gaussian Splatting Attention (GSA) in Section~\ref{sec:gsvit}, and the Gaussian Illumination Field (GIF) in Section~\ref{sec:gif}. Finally, Section~\ref{sec:training_objectives} presents the joint training objective under hybrid supervision.

\begin{figure}[t]
    \centering
    \includegraphics[width=\columnwidth]{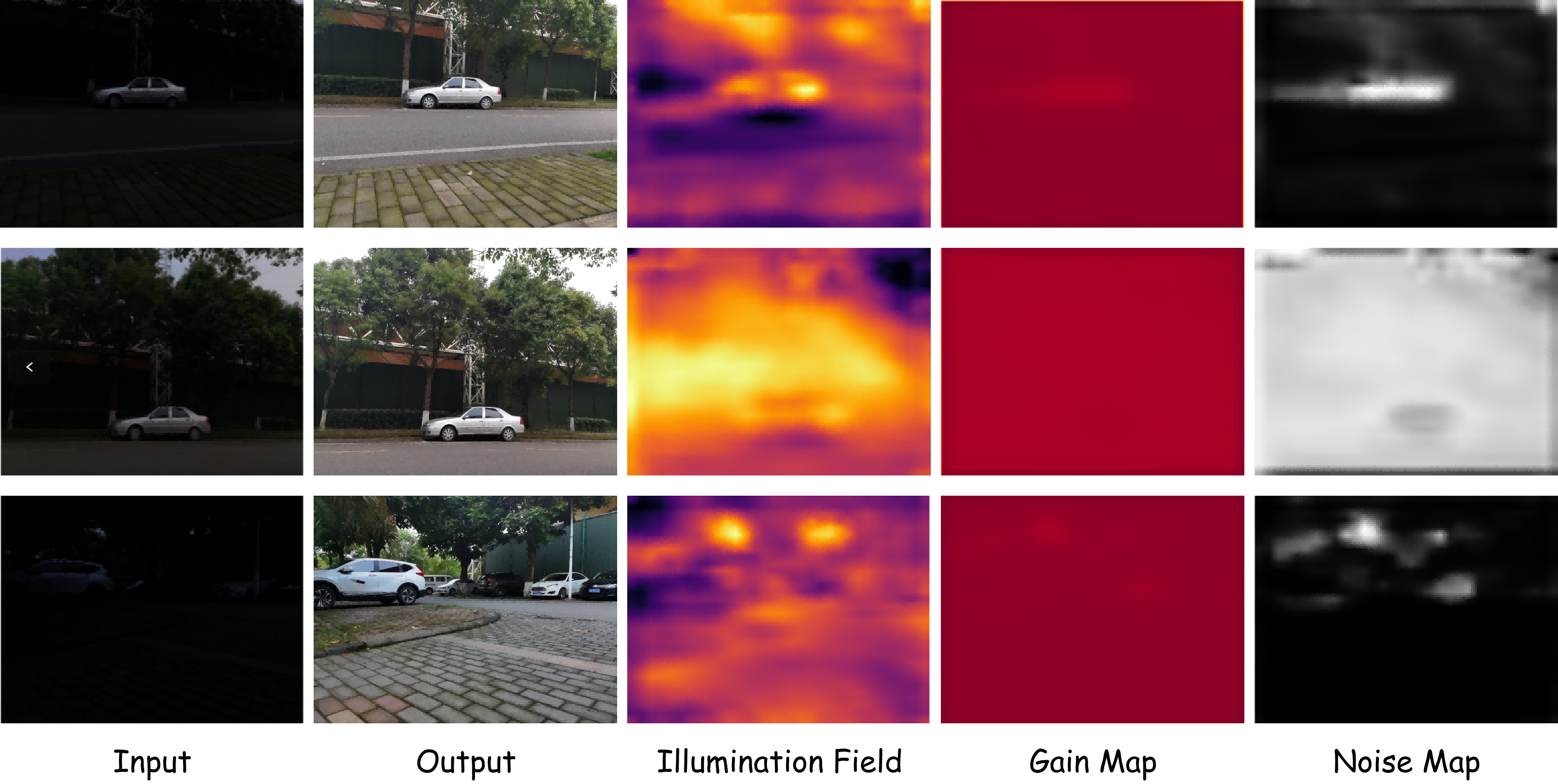}
    \caption{Visualization of physical-prior decomposition and enhancement in GLFS. The outputs corresponding to the components in Eq.~(\ref{eq:enhancement_mapping}) are shown from left to right: the input image, enhanced output, illumination field, spatial gain field, and noise map.}
    \label{fig:physical_decomposition}
\end{figure}

\subsection{Overall Framework}
\label{sec:overall_framework}

The overall training and inference framework of GLFS is illustrated in Fig.~\ref{fig:overall_framework}. The proposed framework adopts an encoder--decoder architecture, with GS-ViT serving as the bottleneck and explicit physical priors embedded into the network. Given a low-light input image $I_l \in \mathbb{R}^{3 \times H \times W}$, the model learns a mapping $\mathcal{G}: I_l \mapsto I_e$ to recover an enhanced image with natural brightness, color fidelity, and structural clarity. During training, we employ a hybrid optimization strategy that combines paired supervision with generative constraints, and jointly optimize the network parameters using reconstruction errors and generative consistency constraints.

To provide the physical motivation for GLFS, we first revisit the light-field degradation mechanism in natural image formation. A low-light image can be modeled as the pixel-wise product of a normal-light image $I_n$ and a spatially varying illumination attenuation field $L_d$, together with additive noise $N_{\mathrm{add}}$:
\begin{equation}
I_l(x)=I_n(x)\odot L_d(x)+N_{\mathrm{add}}(x),
\label{eq:lowlight_degradation}
\end{equation}
where $x \in \Omega \subset \mathbb{R}^{2}$ denotes the pixel coordinate, $\odot$ represents pixel-wise multiplication, $L_d(x)$ denotes the spatially varying illumination attenuation field at pixel $x$, i.e., the illumination degradation component, and $N_{\mathrm{add}}(x)$ represents the additive noise introduced at pixel $x$. Accordingly, the objective of low-light enhancement can be formulated as inferring a spatial gain field $\Gamma(x)$ together with a noise compensation term $\varepsilon(x)$:
\begin{equation}
I_e(x)=I_l(x)\odot \Gamma(x)+\varepsilon(x),
\label{eq:enhancement_mapping}
\end{equation}
where $I_e(x)$ denotes the enhanced output and $\Gamma(x)$ denotes the inferred spatial gain field.

It should be emphasized that Eq.~(\ref{eq:enhancement_mapping}) defines the physical modeling objective of GLFS rather than a pixel-wise operator explicitly hard-coded into the network. In practice, GLFS does not directly apply the multiplication $I_l \odot \Gamma$ to the input. Instead, the continuous gain field $\Gamma(x)$ is injected into the decoder features as a physical conditioning tensor, as described in Eq.~(\ref{eq:gain_injection}). A gated global residual connection in Eq.~(\ref{eq:gated_output}) then forms a convex combination of the generated branch and the input. The noise compensation term $\varepsilon(x)$ is implicitly modeled by the generation branch, while its spatial distribution is conditioned on the noise response map $N(x)$ produced by GIF. This implementation, based on conditional injection and a gated residual, preserves the physical interpretation of Eq.~(\ref{eq:enhancement_mapping}) while avoiding direct noise amplification in dark regions caused by explicit multiplicative enhancement.

The core idea of GLFS is to explicitly represent the latent illumination field $L(x)$, which characterizes the light-field structure and determines the spatial distribution of $\Gamma(x)$, as a continuous superposition of anisotropic two-dimensional Gaussian basis functions. In this way, both the attention computation and the decoding process are constrained by the same set of differentiable geometric parameters. Given $K$ Gaussian basis functions, the light field is analytically formulated as
\begin{equation}
L(x)=\sum_{k=1}^{K}\alpha_k G_k\left(x;\mu_k;\Sigma_k^{-1}\right),
\label{eq:gaussian_light_field}
\end{equation}
where $\mu_k \in \mathbb{R}^{2}$ denotes the center of the $k$-th Gaussian basis function, $\Sigma_k^{-1} \in \mathbb{R}^{2\times2}$ denotes its anisotropic inverse covariance matrix, and $\alpha_k \in [0,1]$ represents its opacity. The Gaussian basis function $G_k\left(x;\mu_k;\Sigma_k^{-1}\right)$ is defined as
\begin{equation}
G_k\left(x;\mu_k;\Sigma_k^{-1}\right)
=
\exp\left(
-\frac{1}{2}
(x-\mu_k)^{\top}
\Sigma_k^{-1}
(x-\mu_k)
\right).
\label{eq:gaussian_basis}
\end{equation}

The above formulation inherits the continuous-field modeling capability of 2DGS. Each Gaussian basis function can be interpreted as a two-dimensional light spot with adjustable principal orientation, spatial extent, and intensity. The superposition of multiple such primitives enables GLFS to characterize complex illumination diffusion and attenuation patterns using a sparse set of differentiable parameters, thereby providing the physical basis for the bias introduced into the subsequent attention mechanism.

We first describe the encoding stage. During inference, the input image can be reflection-padded so that its spatial dimensions are divisible by 8, and the original resolution is restored at the output stage. During training, the input is resized according to the adopted data preprocessing strategy. It is then processed by a $7\times7$ convolution followed by three downsampling blocks with a stride of 2, producing hierarchical features at four different scales:
\begin{equation}
\left\{F_s\right\}_{s=1}^{4}=\mathcal{E}(I_l).
\label{eq:encoder_features}
\end{equation}

These features encode texture, contour, and global context information at different scales. We then perform Gaussian tokenization. The multi-scale Gaussian tokenizer $\Psi_{\mathrm{MGT}}$ infers the Gaussian triplets $(\mu,\Sigma^{-1},\alpha)$ together with the corresponding semantic token tensor $T$ from the two deepest feature levels:
\begin{equation}
(\mu,\Sigma^{-1},\alpha,T)
=
\Psi_{\mathrm{MGT}}(F_4,F_3).
\label{eq:mgt}
\end{equation}
The resulting parameters are subsequently fed into a bottleneck composed of a sequence of GS-ViT blocks. Through repeated non-local interactions with explicit Gaussian biases, the Gaussian parameters and tokens are progressively refined into $(\mu',\Sigma'^{-1},\alpha',T')$, enabling joint modeling of the physical illumination prior and global contextual information.

Finally, the refined representations are passed to the light-field reconstruction and decoding stage. The Gaussian light-field reconstruction module $\Psi_{\mathrm{GIF}}$ splats the refined representations to generate a continuous intrinsic illumination field $L$, a spatial gain field $\Gamma$, and a noise map $N_{\mathrm{map}}$:

\begin{equation}
(L,\Gamma,N_{\mathrm{map}})
=
\Psi_{\mathrm{GIF}}(\alpha',T'),
\label{eq:gif}
\end{equation}
where $\mu'$ and $\Sigma'^{-1}$ have already implicitly conditioned $T'$ through Gaussian-biased attention, while $\alpha'$ directly participates in the estimation of the spatial gain field. GLFS then produces the final enhanced result through a gated global residual together with multi-scale feature fusion in the decoder. The physical interpretability of the overall pipeline is intuitively illustrated in Fig.~\ref{fig:physical_decomposition}, where the low-light input is naturally decomposed into four physically interpretable visual components: the intrinsic illumination field, spatial gain field, noise estimation, and final enhanced output.

\subsection{Multi-Scale Gaussian Tokenizer}
\label{sec:mgt}

\begin{figure}[t]
    \centering
    \includegraphics[width=\columnwidth]{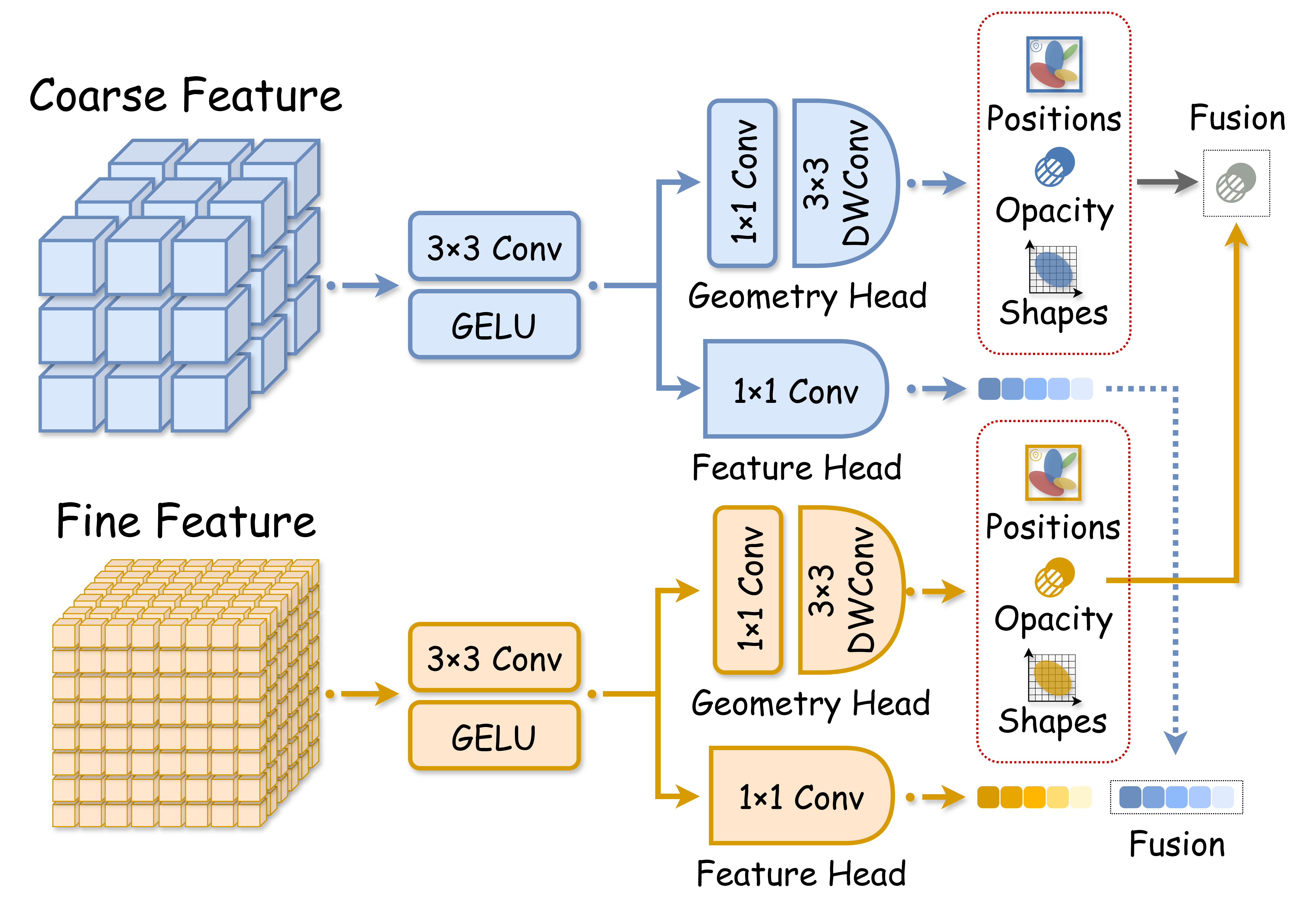}
    \caption{Detailed architecture of the proposed Multi-Scale Gaussian Tokenizer (MGT). The module employs a sequence of $1\times1$ convolutions, $3\times3$ depthwise convolutions, and GELU activations to jointly predict geometric parameters, including position, opacity, and shape, together with semantic features from coarse- and fine-grained feature maps.}
    \label{fig:mgt_architecture}
\end{figure}

Standard ViT tokenization typically partitions an image into regular grid patches, where each token carries positional information but has no explicit physical meaning. Although this design enables non-local interactions among tokens, it cannot directly characterize the geometric structure of scene illumination and forces the subsequent attention mechanism to learn physical priors entirely from data. This limitation becomes a major bottleneck under spatially non-uniform illumination. To address this issue, we propose a Multi-Scale Gaussian Tokenizer (MGT), whose detailed architecture is illustrated in Fig.~\ref{fig:mgt_architecture}. Given the coarse- and fine-grained features produced by the encoder, MGT constructs parallel geometry and feature heads. Specifically, the geometry head employs $1\times1$ convolutions and $3\times3$ depthwise separable convolutions to infer Gaussian geometric parameters, including center positions, opacities, and shapes. In parallel, the feature head extracts semantic features through additional convolutional layers. The resulting geometric parameters and semantic features are then fused across multiple scales, enabling joint multi-scale estimation of the Gaussian physical parameters.

\begin{figure*}[t]
    \centering
    \includegraphics[width=0.75\textwidth]{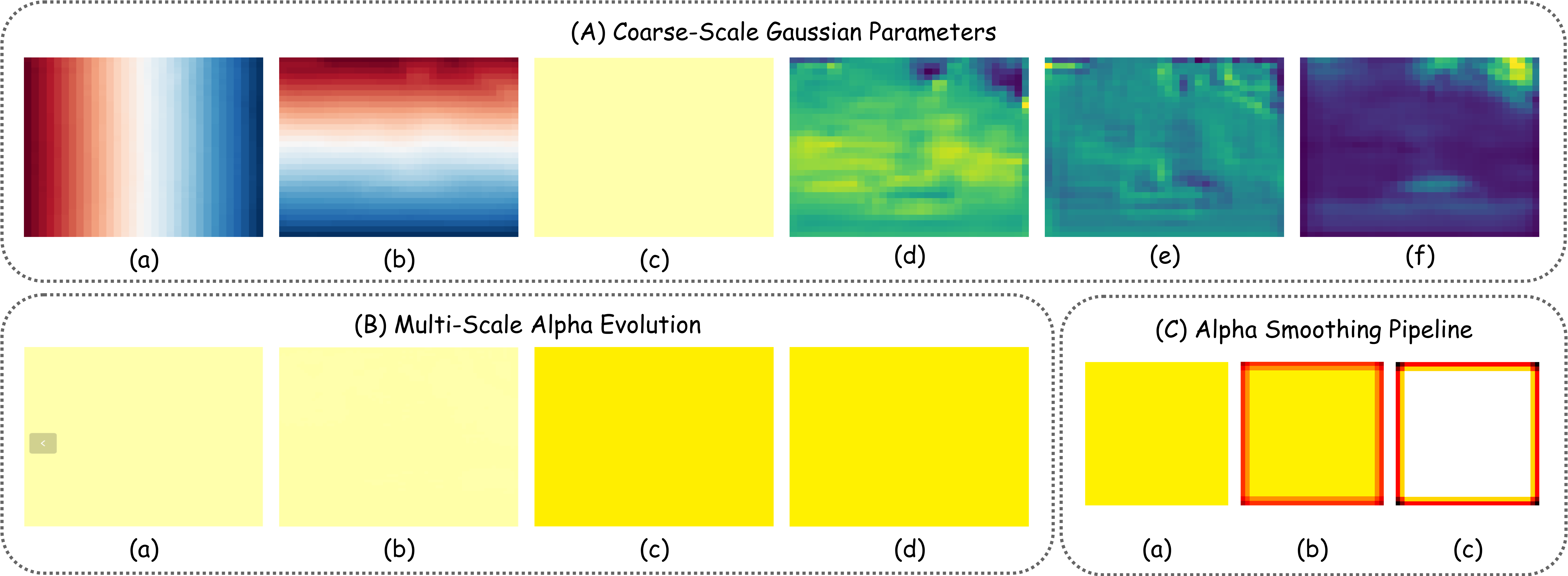}
    \caption{Visualization of the Gaussian parameters inferred by the Multi-Scale Gaussian Tokenizer (MGT). (A) From (a) to (f), the center offsets $(\mu_x,\mu_y)$, opacity $\alpha_c$, and covariance components $(a,b,c)$ predicted by the coarse-grained branch are visualized, revealing the anisotropic geometry of the light field. (B) From (a) to (d), the opacity is visualized at four successive stages---coarse, fine, fused, and refined---illustrating the complementary modeling of global illumination and local details. (C) Comparison of the opacity smoothing process, where abrupt variations between discrete tokens are suppressed to provide continuous geometric guidance for subsequent light-field reconstruction.}
    \label{fig:mgt_parameter_visualization}
\end{figure*}

MGT consists of a coarse-grained branch and a fine-grained branch, which operate on the two deepest encoder features, $F_4$ and $F_3$, respectively. The two branches are designed to collaboratively characterize physical signals at different scales, namely global illumination distribution and local high-frequency compensation. In each branch, a shared $3\times3$ convolutional backbone is first employed to extract spatial contextual information. Two parallel $1\times1$ convolutional heads are then used to predict a 6-dimensional geometric descriptor tensor and a $d$-dimensional semantic feature tensor, respectively. The six geometric channels correspond to the center offsets (2 dimensions), the independent elements of the inverse covariance matrix (3 dimensions), and the opacity (1 dimension). Before output, the predicted 6-dimensional geometric tensor is further processed by a $3\times3$ depthwise smoothing convolution whose weights are initialized with an averaging kernel. This operation suppresses abrupt variations in geometric parameters between neighboring tokens and provides continuous geometric guidance for subsequent light-field reconstruction, as illustrated in Fig.~\ref{fig:mgt_parameter_visualization}(C).

To constrain the center position $\mu$ to the normalized pixel-coordinate domain $[-1,1]^2$, we anchor it to a fixed normalized coordinate grid $P$ and let the network predict a small relative offset $\Delta_{\mu}$, which is bounded by a hyperbolic tangent function and scaled according to the feature-map resolution:
\begin{equation}
\mu(x)
=
P(x)
+
\tanh\left(\Delta_{\mu}(x)\right)
\cdot
\frac{2}{\min\left(H',W'\right)},
\label{eq:gaussian_center}
\end{equation}
where $H'$ and $W'$ denote the spatial resolution of the coarse-grained feature map. This parameterization anchors each Gaussian basis function around its corresponding grid location, thereby reducing the training instability associated with unconstrained coordinate prediction.

To ensure a valid anisotropic inverse covariance matrix,
$\Sigma^{-1}=
\begin{bmatrix}
a & b\\
b & c
\end{bmatrix}$,
positive definiteness must be guaranteed. For the diagonal elements, we adopt a Softplus parameterization with an additive bias of 1 and an upper bound to ensure non-degeneracy and boundedness:
\begin{equation}
a(x)
=
\min\left(
\operatorname{Softplus}\left(\widetilde{a}(x)\right)+1,
a_{\max}
\right),
\label{eq:covariance_a}
\end{equation}
where $a_{\max}=200$. The parameter $c(x)$ follows the same formulation as $a(x)$, with the two quantities predicted from independent channels of the geometric tensor, i.e., $\widetilde{a}$ and $\widetilde{c}$, respectively. The additive bias of 1 ensures that the diagonal elements $a(x)$ and $c(x)$ are no smaller than 1, preventing the diagonal precision terms from approaching zero and the Gaussian support from becoming excessively broad. The upper bound $a_{\max}$ prevents numerical divergence of the inverse covariance matrix during training. The off-diagonal element is parameterized as

\begin{figure*}[t]
    \centering
    \includegraphics[width=0.75\textwidth]{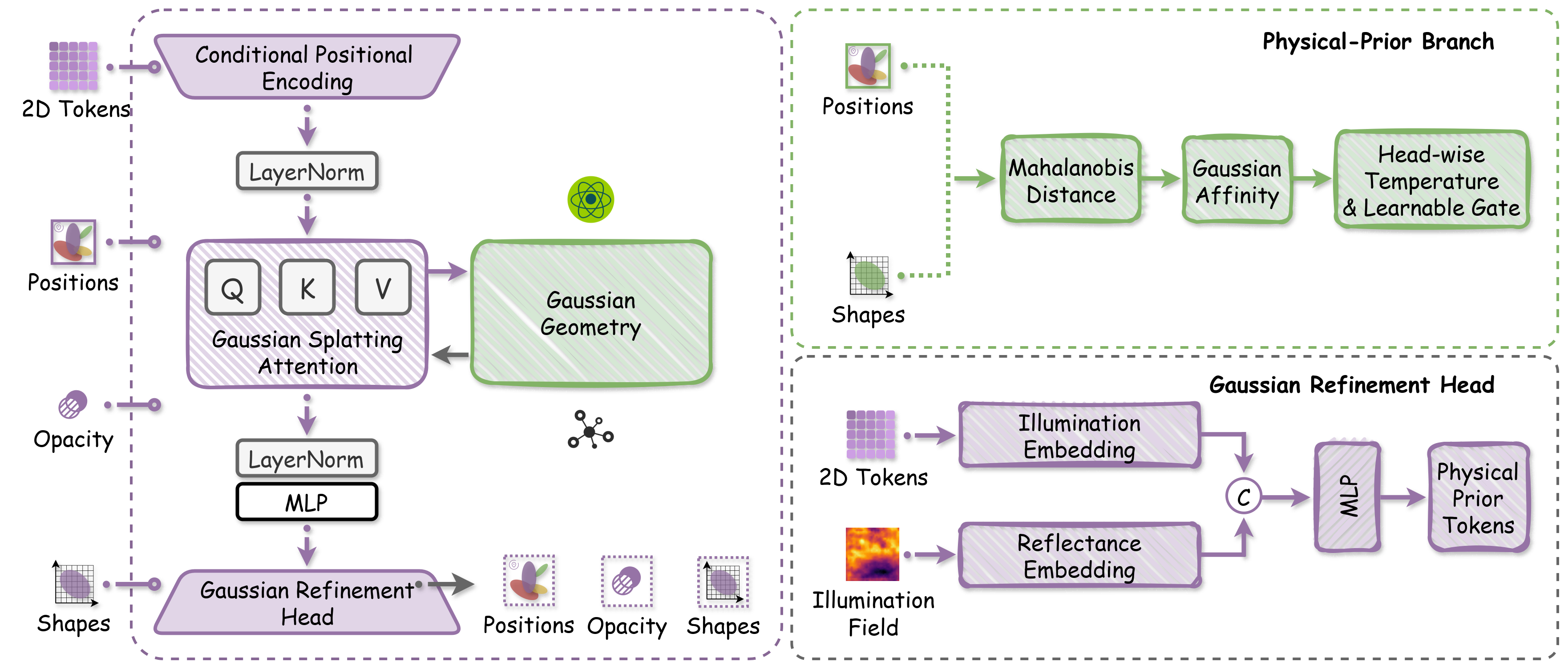}
    \caption{Detailed architecture of the proposed GS-ViT module. In addition to standard conditional positional encoding and self-attention, GS-ViT explicitly introduces a physical-prior branch and illustrates the internal data flow of the pairwise Gaussian affinity module used to generate the physical affinity matrix.}
    \label{fig:gsvit_architecture}
\end{figure*}

\begin{equation}
b(x)
=
\tanh\left(\widetilde{b}(x)\right)
\cdot
\frac{1}{2}
\sqrt{a(x)c(x)}.
\label{eq:covariance_b}
\end{equation}
This parameterization is geometrically equivalent to the explicit LDL decomposition used in 3DGS, while being more suitable for batched computation over dense convolutional outputs. The opacity $\alpha$ is generated using a Sigmoid function with a learnable bias $b_{\alpha}$:
\begin{equation}
\alpha(x)
=
\sigma\left(\widetilde{\alpha}(x)+b_{\alpha}\right),
\label{eq:opacity_parameterization}
\end{equation}
where $b_{\alpha}$ is initialized to a positive value to encourage dense activation during the early stage of training, thereby accelerating the convergence of Gaussian geometry in MGT.

To jointly characterize global illumination and local details, the coarse- and fine-grained branches of MGT produce $(\mu_c,\Sigma_c^{-1},\alpha_c,T_c)$ and $(\mu_f,\Sigma_f^{-1},\alpha_f,T_f)$, respectively. The semantic tokens from the fine-grained branch are first downsampled using a stride-2 convolution to match the spatial resolution of the coarse-grained branch. They are then concatenated with the coarse-grained semantic tokens and fused through a $1\times1$ convolution $\Phi_T$ to obtain the final Gaussian token representation:

\begin{equation}
T
=
\Phi_T\left(
T_c \parallel \operatorname{Down}(T_f)
\right),
\label{eq:token_fusion}
\end{equation}
where $\parallel$ denotes channel-wise concatenation and $\operatorname{Down}_{T}(\cdot)$ denotes stride-2 convolutional downsampling. Multi-scale opacity fusion is performed using a $2\!\rightarrow\!1$ convolution followed by a Sigmoid function, allowing opacity information from the coarse- and fine-grained branches to be adaptively weighted:

\begin{equation}
\alpha
=
\sigma\left(
\Phi_{\alpha}\left(
\alpha_c \parallel \operatorname{Down}(\alpha_f)
\right)
\right),
\label{eq:opacity_fusion}
\end{equation}
where $\operatorname{Down}_{\alpha}(\cdot)$ denotes bilinear resampling of the fine-grained opacity map to the coarse-grained resolution. The spatial geometry of the light field is primarily determined by the global illumination distribution, and the consistency of geometric parameters across the token sequence directly affects the numerical stability of the attention bias introduced in Section~\ref{sec:gsvit}. Therefore, the geometric parameters passed to the bottleneck are taken from the coarse-grained branch, i.e., $\mu=\mu_c$ and $\Sigma^{-1}=\Sigma_c^{-1}$. The fine-grained branch contributes through the fused opacity $\alpha$ and semantic tokens $T$. Through this multi-scale coordination, the global light-field geometry is determined by the coarse-grained parameters $(\mu_c,\Sigma_c^{-1})$, while $\alpha_c$ provides the coarse-scale opacity response and $\alpha_f$ complements local details such as reflections and shadow boundaries. These multi-scale opacity cues are jointly represented in the fused $\alpha$. This process is illustrated in Fig.~\ref{fig:mgt_parameter_visualization}.

\begin{figure}[t]
    \centering
    \includegraphics[width=0.9\columnwidth]{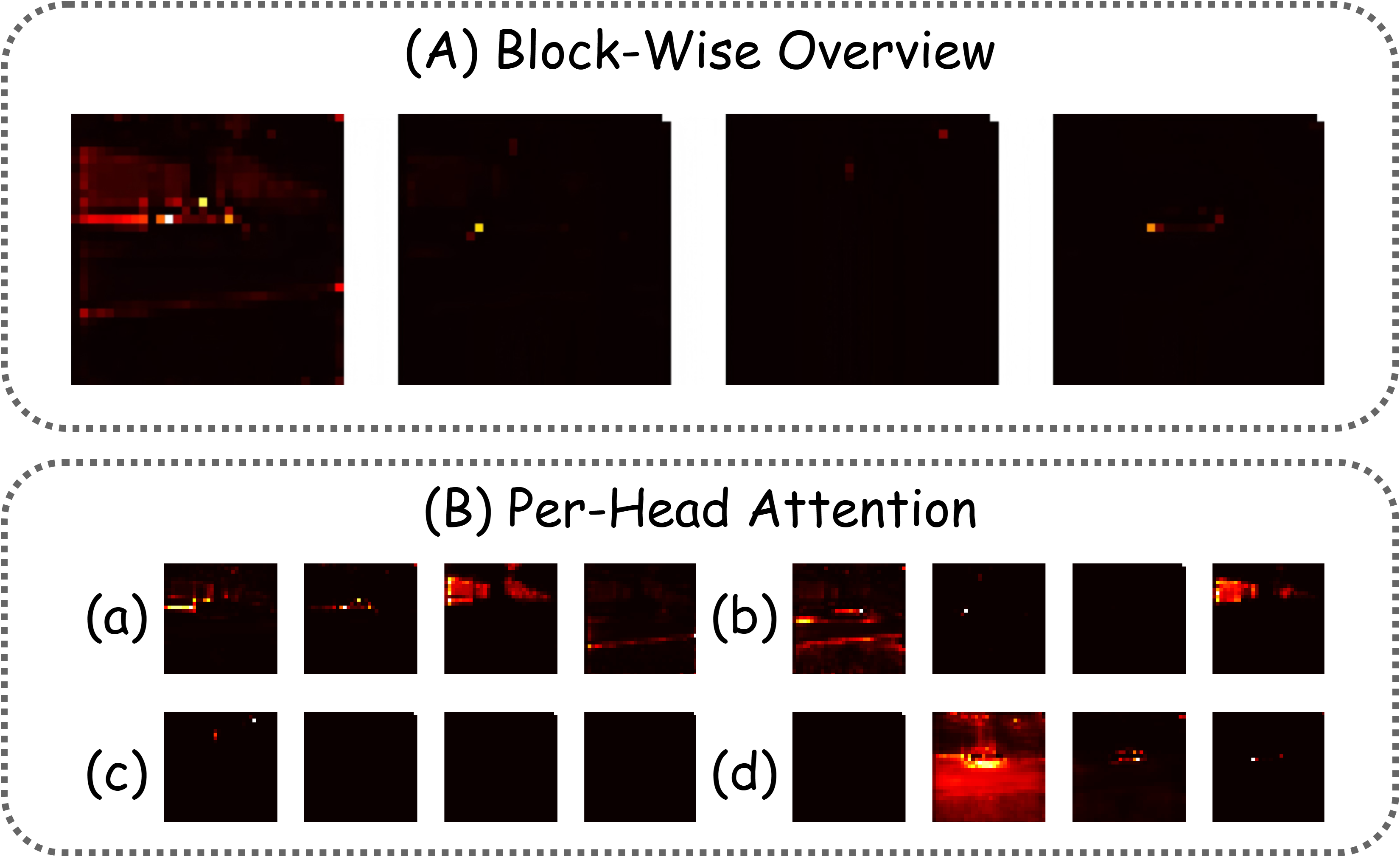}
    \caption{Hierarchical evolution of anisotropic Gaussian splatting attention in GS-ViT. (A) Attention distributions of the central token across four GS-ViT blocks. The attention is relatively diffuse in shallow blocks and progressively evolves into anisotropic structures centered around the light source in deeper blocks, showing high consistency with the opacity distribution. (B) Independent responses of different attention heads. Through the learnable temperature $t_h$, different heads collaboratively model the light field at different spatial scales and orientations.}
    \label{fig:gsvit_attention_evolution}
\end{figure}

\subsection{GS-ViT with Gaussian Splatting Attention}
\label{sec:gsvit}

Standard ViTs perform non-local modeling through scaled dot-product attention. However, their relative positional biases are typically represented by either learnable parameters or coordinate differences and are thus decoupled from the physical geometry of scene illumination. Under low-light conditions, this prevents pixels illuminated by the same physical light source from automatically establishing strong connections based on physical affinity, which can lead to granular artifacts under spatially non-uniform illumination. To address this issue, we explicitly inject an anisotropic physical bias driven by Gaussian geometry into the GS-ViT block, whose detailed architecture is illustrated in Fig.~\ref{fig:gsvit_architecture}. In addition to the standard attention computation based on query ($Q$), key ($K$), and value ($V$), GS-ViT introduces a dedicated physical-prior branch. This branch concatenates geometric distance, appearance similarity, and shape consistency along the channel dimension and then employs a multilayer perceptron (MLP) to generate a physical affinity matrix. The resulting affinity guides the computation of self-attention scores and the refinement of Gaussian parameters, thereby constraining long-range interactions according to the underlying light-field structure.

Given an input token sequence $T\in\mathbb{R}^{B\times N\times d}$, where $N_{\mathrm{token}}=H'W'$, we first inject spatial structural cues through Conditional Position Encoding (CPE) implemented using a depthwise convolution, thereby avoiding the limitations of hard-coded positional embeddings for varying input resolutions:
\begin{equation}
T
\leftarrow
T+
\operatorname{Reshape}\left(
\operatorname{DWConv}_{3\times3}
\left(
\operatorname{Reshape}(T)
\right)
\right).
\label{eq:cpe}
\end{equation}
Standard multi-head linear projections are then applied to obtain the query, key, and value tensors:
\begin{equation}
[Q,K,V]
=
\operatorname{Reshape}\left(TW_{qkv}\right)
\in
\mathbb{R}^{B\times h\times N\times d_h},
\label{eq:qkv_projection}
\end{equation}
where $h$ denotes the number of attention heads and $d_h=d/h$ is the dimension of each head.

The main modification to the attention score computation is the incorporation of the predicted Gaussian geometry as an anisotropic physical bias. For an arbitrary token pair $(i,j)$, we define their relative position difference in the normalized coordinate system as $\Delta x_{ij}=\mu_i-\mu_j$ and use the inverse covariance matrix associated with the query token $i$ as the metric. The corresponding anisotropic squared distance is defined as
\begin{equation}
d_{ij}^{2}
=
a_i\Delta x_{ij,1}^{2}
+
2b_i\Delta x_{ij,1}\Delta x_{ij,2}
+
c_i\Delta x_{ij,2}^{2}.
\label{eq:mahalanobis_distance}
\end{equation}
This distance directly inherits the anisotropic characteristics of the Gaussian basis functions. When $\Sigma_i^{-1}$ has a larger eigenvalue along a particular direction, displacements along that direction contribute more strongly to the distance, whereas those along directions with smaller eigenvalues contribute less. We then map this distance to a bounded Gaussian affinity:
\begin{equation}
A_{ij}
=
\exp\left(
-\frac{1}{2}d_{ij}^{2}
\right).
\label{eq:gaussian_affinity}
\end{equation}

To allow different attention heads to flexibly control the scale of the physical bias, each head is assigned an independent learnable temperature $t_h$. Its positivity is enforced through Softplus, while truncation is applied to prevent numerical overflow:
\begin{equation}
t_h
=
\min\left(
\operatorname{Softplus}\left(\widetilde{t}_h\right),
t_{\max}
\right),
\qquad
t_{\max}=2.
\label{eq:temperature_parameterization}
\end{equation}

The final attention logits are obtained by combining the semantic dot-product term with the physical Gaussian bias through a learnable scalar gate $\beta=\sigma(\widetilde{\beta})$:
\begin{equation}
\operatorname{Logit}_h(i,j)
=
\frac{Q_{h,i}K_{h,j}^{\top}}{\sqrt{d_h}}
+
\beta \cdot t_h \cdot A_{ij}.
\label{eq:attention_logit}
\end{equation}
Softmax normalization is then applied along the $j$ dimension, followed by multiplication with $V$. The outputs of all attention heads are concatenated and linearly projected back to the $d$-dimensional feature space:
\begin{equation}
\operatorname{Out}
=
\operatorname{Concat}_{h}
\left(
\operatorname{Softmax}_{j}
\left(
\operatorname{Logit}_{h}(i,j)
\right)
V_h
\right)
W_o.
\label{eq:attention_output}
\end{equation}

To balance numerical stability and memory efficiency, the squared Mahalanobis distance in Eq.~(\ref{eq:mahalanobis_distance}) is clamped to $[0,20]$ during implementation, since Gaussian affinities beyond the upper bound decay below $e^{-10}$ and are therefore negligible. Moreover, all computations involving Gaussian geometry in Eqs.~(\ref{eq:mahalanobis_distance})--(\ref{eq:attention_logit}) are performed in single precision to improve the numerical stability of covariance-related computations under automatic mixed-precision (AMP) training. The gating parameter is initialized as $\widetilde{\beta}=-1$, corresponding to $\beta\approx0.27$, such that the physical bias exerts only a weak influence on the attention distribution during the early stage of training.

\begin{figure*}[t]
    \centering
    \includegraphics[width=0.75\textwidth]{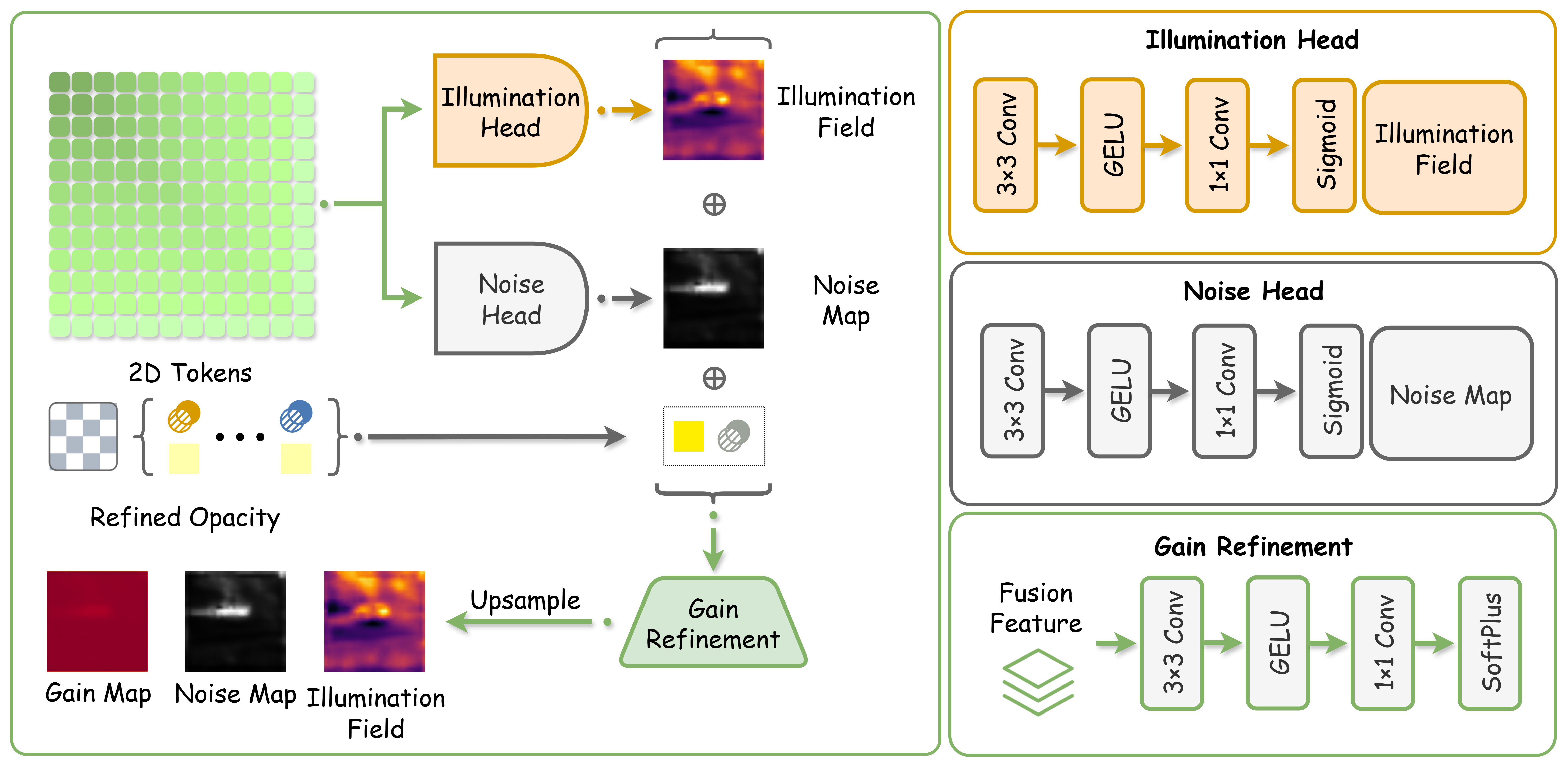}
    \caption{Detailed network configuration of the Gaussian Illumination Field (GIF). The figure illustrates the specific arrangement of convolutional layers and activation functions used to physically decompose the scene into an illumination map, a noise map, and a gain map.}
    \label{fig:gif_architecture}
\end{figure*}

This design provides a clear physical interpretation while addressing the optimization conflict and numerical instability associated with integrating 2DGS into self-attention. When $t_h$ is large, the physical bias has a stronger influence on the attention distribution, encouraging strong connections among tokens within the same light-field support region. Meanwhile, the gate $\beta$ allows the model to rely primarily on semantic similarity during the early stage of training and adaptively adjust the contribution of the physical prior thereafter, reducing the optimization difficulty caused by rigid constraints. The hierarchical evolution of the attention maps is illustrated in Fig.~\ref{fig:gsvit_attention_evolution}. The attention distributions in shallow GS-ViT blocks more closely resemble semantic clustering, whereas those in deeper blocks gradually exhibit anisotropic structures that are consistent with the opacity distribution $\alpha$ and spread outward from the light-source centers.

To enable the Gaussian geometry and semantic features to evolve collaboratively, each GS-ViT block is further equipped with a lightweight geometric refinement branch after the feed-forward layer. Taking the current token features as input, this branch regresses a 6-dimensional residual $[\delta_{\mu},\delta_{\Sigma},\delta_{\alpha}]$ and employs a learnable gate $g=0.3\tanh(\widetilde{g})$ to apply small refinements to the Gaussian parameters. The center position is refined using the same normalized truncation strategy as in Section~\ref{sec:mgt}:
\begin{equation}
\mu'
=
\operatorname{clip}\left(
\mu
+
g\cdot0.01\tanh(\delta_{\mu}),
-1,1
\right).
\label{eq:gaussian_center_refinement}
\end{equation}
For the inverse covariance matrix, a positive-definite projection $\operatorname{PD}(\cdot)$ is applied after the residual update:
\begin{equation}
\Sigma'^{-1}
=
\operatorname{PD}\left(
\Sigma^{-1}
+
g\cdot0.005\,\delta_{\Sigma}
\right).
\label{eq:covariance_refinement}
\end{equation}
The refined opacity is clipped to the valid probability range:
\begin{equation}
\alpha'
=
\operatorname{clip}\left(
\alpha
+
g\cdot0.02\tanh(\delta_{\alpha}),
0,1
\right).
\label{eq:opacity_refinement}
\end{equation}
The residual updates are scaled by small coefficients ranging from $0.005$ to $0.02$. Together with the bounded output of the learnable gate, this design allows the Gaussian geometry to continuously incorporate semantic evidence during layer-wise non-local interactions while keeping the refinements small to preserve the stability of the overall light-field structure. The comparison between the refined opacity $\alpha'$ and the initial opacity $\alpha$ is shown in Fig.~\ref{fig:mgt_parameter_visualization}(C). After refinement through the GS-ViT bottleneck, the opacity response becomes sharper around boundaries while remaining smoother within regions covered by spatially uniform illumination.

\subsection{Gaussian Illumination Field}
\label{sec:gif}

After refinement through the GS-ViT bottleneck, the geometric parameters of each token are aligned with the global semantic information. However, the Gaussian parameters themselves still constitute a discrete and sparse representation. To transform this sparse representation into a continuous light field that can be fused pixel-wise with the decoder features, we design a Gaussian light-field reconstruction (GIF) module, whose detailed layer-wise architecture is shown in Fig.~\ref{fig:gif_architecture}. GIF consists of three parallel prediction heads. The illumination prediction head and the noise estimation head adopt the same network architecture and are used to reconstruct the intrinsic illumination field and noise map, respectively. The gain refinement head employs a Softplus activation function to map the fused features to physically plausible positive gain values.

\noindent\textbf{(i) Intrinsic Illumination Prediction Head $\boldsymbol{\Phi}_{L}$:}
This head takes the refined token tensor $T'$, reshaped into its two-dimensional spatial form, as input. A two-layer convolutional network consisting of a $3\times3$ convolution followed by a $1\times1$ convolution and a Sigmoid activation is used to predict an intrinsic illumination map with a spatial resolution of $H'\times W'$:
\begin{equation}
L(x)=\sigma\left(\Phi_{L}\left(T'(x)\right)\right),
\label{eq:intrinsic_illumination}
\end{equation}
where $L(x)\in[0,1]$ represents the normalized luminous flux received at pixel $x$. Its spatial distribution corresponds to the splatting-based superposition of Gaussian basis functions described in Section~\ref{sec:overall_framework}. A low value indicates a severely attenuated dark region, whereas a high value corresponds to a well-illuminated region.

\noindent\textbf{(ii) Noise Estimation Head $\boldsymbol{\Phi}_{N}$:}
The noise level in low-light images is often negatively correlated with local brightness and exhibits complex spatial correlations. Using the same refined token tensor $T'$ as input, this head independently predicts a normalized noise response map:

\begin{equation}
N(x)=\sigma\left(\Phi_{N}\left(T'(x)\right)\right),
\label{eq:noise_response}
\end{equation}
where $N(x)$ serves as one of the conditioning inputs to the gain refinement head in Eq.~(\ref{eq:coarse_gain_field}), enabling the gain estimation to explicitly account for the local noise level and thereby suppress excessive amplification in dark regions. Meanwhile, it provides an interpretable characterization of the spatial distribution associated with the noise compensation term $\varepsilon(x)$ in Eq.~(\ref{eq:enhancement_mapping}), as illustrated in Fig.~\ref{fig:physical_decomposition}. No additional explicit supervision or regularization is imposed on $N(x)$; instead, its response is learned implicitly through downstream gradients from the gain field.

\noindent\textbf{(iii) Gain Field Refinement Head $\Phi_\Gamma$:}
We concatenate the illumination field $L$, noise response map $N$, and refined opacity $\alpha'$ along the channel dimension to form a three-channel conditioning tensor. A coarse gain field is then predicted using two convolutional layers followed by a $\operatorname{Softplus}$ activation. An additive bias of $1$ shifts the predicted gain to the amplification regime above $1$:
\begin{equation}
\Gamma_{\mathrm{low}}(x)
=
1+\operatorname{Softplus}\!\left(
\Phi_\Gamma
\left(
L(x)\parallel N(x)\parallel\alpha'(x)
\right)
\right).
\label{eq:coarse_gain_field}
\end{equation}

To suppress block artifacts caused by local discontinuities in the Gaussian geometry, we first bilinearly upsample $\Gamma_{\mathrm{low}}$ to the full spatial resolution. The upsampled gain field is then smoothed using a fixed $5\times5$ isotropic Gaussian kernel $K_{\sigma}$ with $\sigma=1$, whose parameters are not learnable. Finally, the smoothed gain field is clipped to a physically reasonable amplification range:
\begin{equation}
\Gamma(x)
=
\operatorname{clip}\!\left(
K_{\sigma}\ast
\operatorname{Up}\!\left(\Gamma_{\mathrm{low}}\right)(x),
\Gamma_{\min},
\Gamma_{\max}
\right),
\end{equation}
where $\Gamma_{\min}=1$ ensures that the gain field does not reduce the input brightness, i.e., it only enhances rather than darkens the input, while $\Gamma_{\max}=4$ is determined empirically to provide a reasonable balance between visibility improvement and noise amplification. This smoothing-and-clipping strategy suppresses token-boundary discontinuities and prevents local overexposure.

The final enhanced image is synthesized through two complementary paths. The first path uses the decoder feature $D$, which is obtained through multi-scale upsampling and channel-attention-guided fusion with the encoder features. A gain-injection convolution $\Phi_{\mathrm{inj}}$ is then used to align and fuse the downsampled gain field with the decoder feature:
\begin{equation}
\widetilde{D}
=
\Phi_{\mathrm{inj}}
\left(
D \parallel \operatorname{Down}(\Gamma)
\right).
\label{eq:gain_injection}
\end{equation}
The second path is a direct global residual connection, where a learnable gate $\omega=\sigma(\check{\omega})$ controls the mixing ratio between the generated output and the input:

\begin{equation}
I_e
=
(1-\omega)
\tanh\left(
\Phi_{\mathrm{out}}(\widetilde{D})
\right)
+
\omega I_l,
\label{eq:gated_output}
\end{equation}
where $\Phi_{\mathrm{out}}$ denotes the output head, which consists of reflection padding followed by a $7\times7$ convolution. The gate $\omega$ is initialized to a small Sigmoid response, i.e., $\sigma(-1.5)\approx0.18$. This gated residual design preserves the global structure while allowing the generation path to perform local gain and color correction. As illustrated in Fig.~\ref{fig:physical_decomposition}, the low-light input is decomposed into three physically interpretable visual components, namely the intrinsic illumination field, spatial gain field, and noise response map, which are ultimately combined to produce an enhanced image with uniform illumination and faithful colors.

\subsection{Joint Training Objectives}
\label{sec:training_objectives}
\begin{figure*}[t]
    \centering
    \includegraphics[width=0.9\textwidth]{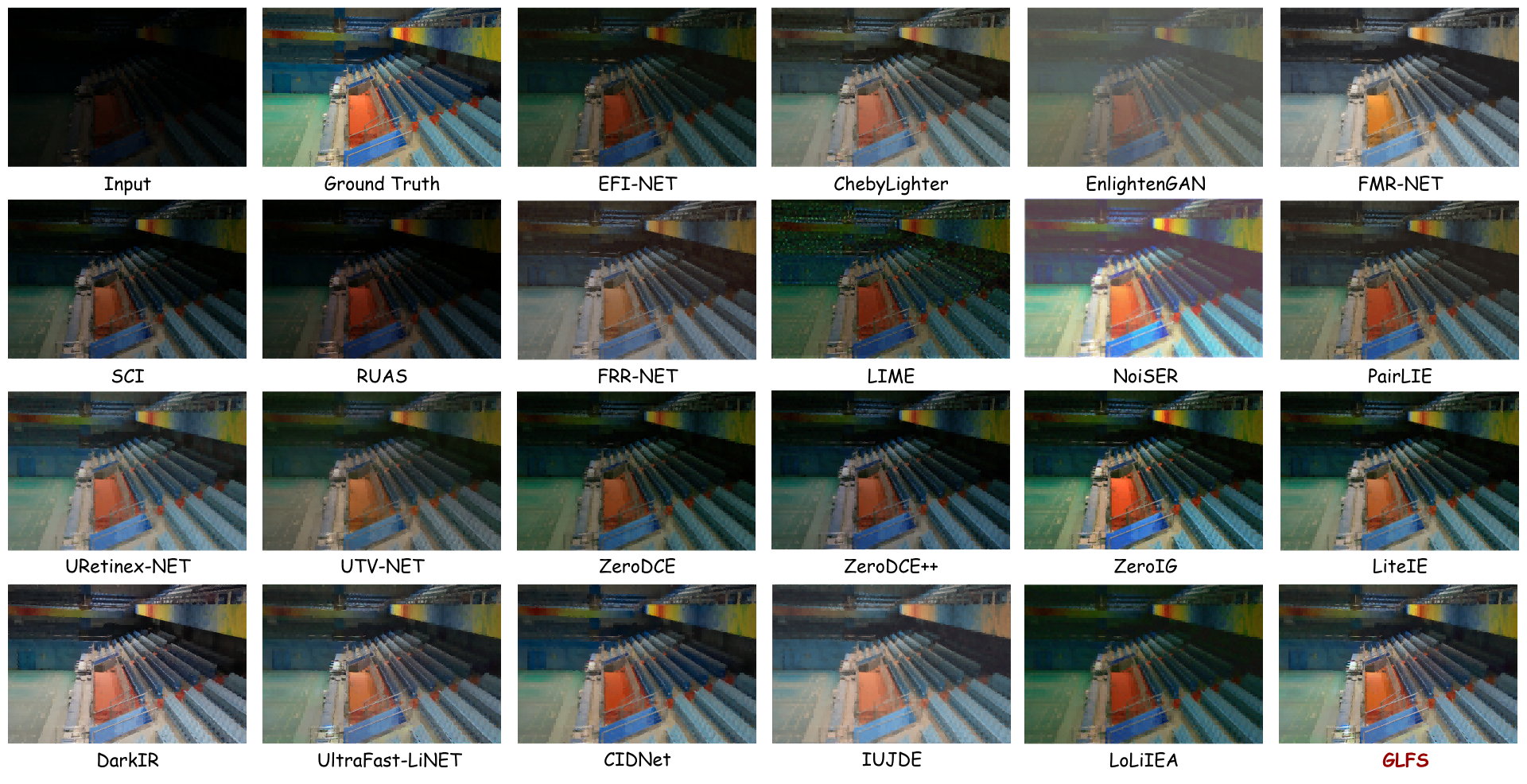}
    \caption{Visual comparison of GLFS with SOTA methods on the LOL-V1 dataset.}
    \label{fig:compare_lol_v1}
\end{figure*}

\begin{table*}[t]
\centering
\renewcommand{\arraystretch}{0.85}
\caption{Performance comparison on the LOL-v1 dataset. The best and second-best results for each metric are highlighted in red and blue, respectively.}
\label{tab:performance_lolv1}
\resizebox{0.95\textwidth}{!}{
\begin{tabular}{l|cccccccccc}
\toprule
Method & SSIM$\uparrow$ & PSNR$\uparrow$ & LPIPS$\downarrow$ & NIQE$\downarrow$ & LOE$\downarrow$ & DE$\uparrow$ & EME$\uparrow$ & Params(M)$\downarrow$ & FLOPs(G)$\downarrow$ & Time(ms)$\downarrow$ \\
\midrule
FMR-NET~\cite{ref11}       & 0.79 & 18.87 & 0.27 & 4.63 & 26.00 & 2.29 & 2.97 & 0.197 & 205.658 & 8.74 \\
FRR-NET~\cite{ref12}       & 0.75 & 18.28 & 0.29 & 5.79 & 24.87 & 1.95 & 3.83 & 0.0183 & 0.665 & 5.77 \\
UTV-NET~\cite{ref13}       & 0.74 & 15.62 & 0.22 & 4.31 & 29.10 & 2.04 & 4.57 & 7.751 & 41.671 & 31.78 \\
EFI-NET~\cite{ref15}       & 0.64 & 13.48 & 0.31 & 4.43 & 24.18 & 1.65 & 5.35 & 0.129 & 18.761 & 6.434 \\
URetinex-NET~\cite{ref18}  & 0.83 & 20.50 & 0.13 & \best{4.06} & 29.75 & 2.13 & 3.45 & 0.340 & 79.141 & 4.726 \\
DarkIR~\cite{ref63}        & \second{0.84} & \second{24.17} & \second{0.10} & 4.61 & 23.58 & 2.37 & 3.70 & 3.322 & 14.311 & 17.366 \\
CIDNet~\cite{ref64}        & \best{0.86} & 24.12 & \best{0.09} & 4.22 & 38.20 & 2.38 & 3.84 & 1.976 & 16.412 & 23.303 \\
LoLi-IEA~\cite{ref68}      & 0.59 & 17.12 & 0.33 & 6.68 & 37.81 & 2.11 & 6.99 & 26.593 & 58.360 & 4.347 \\
LiteIE~\cite{ref4}         & 0.62 & 19.21 & 0.31 & 5.90 & \second{3.10} & 2.32 & 19.29 & \best{0.000058} & 0.0259 & \second{1.214} \\
LIME~\cite{ref5}           & 0.50 & 14.01 & 0.42 & 5.61 & 102.02 & 1.84 & \best{22.26} & -- & -- & 858.343 \\
ZeroDCE++~\cite{ref6}      & 0.61 & 15.38 & 0.32 & 5.86 & 25.81 & 1.99 & 19.57 & 0.011 & \second{0.024} & 1.806 \\
ZeroDCE~\cite{ref7}        & 0.60 & 14.99 & 0.32 & 5.81 & 26.39 & 1.84 & 19.54 & 0.079 & 10.384 & 1.604 \\
SCI~\cite{ref8}            & 0.56 & 13.87 & 0.34 & 6.00 & 8.43 & 1.92 & 20.18 & \second{0.00026} & 0.032 & \best{0.928} \\
RUAS~\cite{ref9}           & 0.46 & 11.33 & 0.38 & 5.76 & \best{0.23} & 1.47 & \second{20.84} & 0.00344 & 0.441 & 5.108 \\
EnlightenGAN~\cite{ref10}  & 0.72 & 17.06 & 0.30 & 4.22 & 66.68 & 1.65 & 3.05 & 8.637 & 32.783 & 4.212 \\
ChebyLighter~\cite{ref14}  & 0.78 & 19.86 & 0.19 & 4.36 & 16.99 & 2.16 & 3.59 & 0.178 & 22.957 & 19.73 \\
PairLIE~\cite{ref16}       & 0.77 & 18.59 & 0.23 & \second{4.08} & 51.86 & 1.91 & 3.76 & 0.342 & 44.695 & 2.589 \\
NoiSER~\cite{ref17}        & 0.70 & 17.36 & 0.36 & 4.38 & 49.77 & 2.36 & 2.56 & 0.00176 & \best{0.0226} & 1.283 \\
ZeroIG~\cite{ref19}        & 0.50 & 17.98 & 0.37 & 6.45 & 13.45 & \best{2.56} & 20.53 & 0.087 & 20.963 & 1.951 \\
IUJDE~\cite{ref65}         & 0.78 & 18.85 & 0.26 & 4.47 & 30.28 & 2.14 & 2.73 & 0.345 & 46.555 & 24.617 \\
\textbf{GLFS (Ours)}       & \second{0.84} & \best{24.38} & 0.12 & 4.14 & 27.16 & \second{2.44} & 7.59 & 7.804 & 66.824 & 14.471 \\
\bottomrule
\end{tabular}}
\end{table*}

\begin{figure*}
    \centering
    \includegraphics[width=0.9\textwidth]{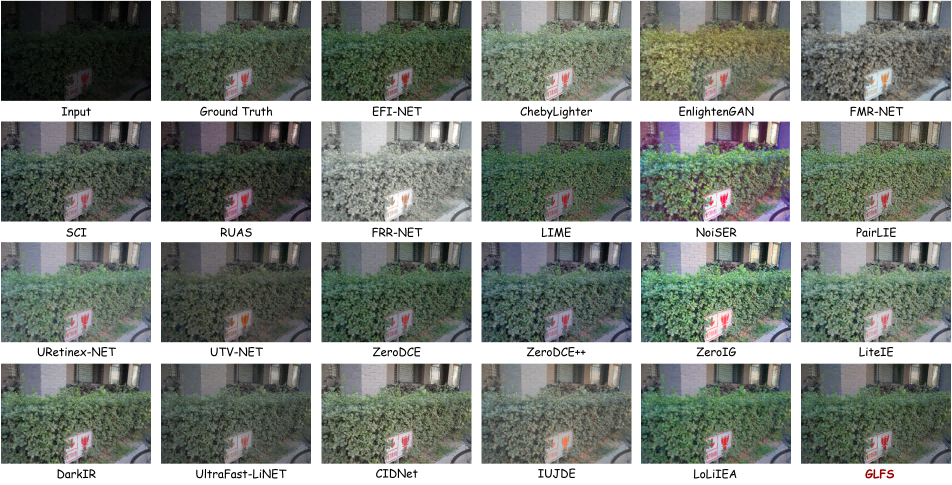}
    \caption{Visual comparison of GLFS with SOTA methods on the LOL-V2-Real dataset.}
    \label{fig:compare_lol_v2_real}
\end{figure*}

\begin{table*}[t]
\centering
\renewcommand{\arraystretch}{0.85}
\caption{Performance comparison on the LOL-v2-Real dataset. The best and second-best results for each metric are highlighted in red and blue, respectively.}
\label{tab:performance_lolv2real}
\resizebox{0.95\textwidth}{!}{
\begin{tabular}{l|cccccccccc}
\toprule
Method & SSIM$\uparrow$ & PSNR$\uparrow$ & LPIPS$\downarrow$ & NIQE$\downarrow$ & LOE$\downarrow$ & DE$\uparrow$ & EME$\uparrow$ & Params(M)$\downarrow$ & FLOPs(G)$\downarrow$ & Time(ms)$\downarrow$ \\
\midrule
FMR-NET~\cite{ref11}       & 0.77 & 19.76 & 0.44 & 5.48 & 76.70 & \second{2.48} & 3.45 & 0.197 & 205.658 & 8.74 \\
FRR-NET~\cite{ref12}       & 0.74 & 17.29 & 0.37 & 4.15 & 69.61 & 2.44 & 4.08 & 0.0183 & 0.665 & 5.77 \\
UTV-NET~\cite{ref13}       & 0.79 & 19.91 & 0.18 & 5.44 & 29.30 & 2.02 & 5.80 & 7.751 & 41.671 & 31.78 \\
EFI-NET~\cite{ref15}       & 0.67 & 17.05 & 0.27 & 4.79 & 33.56 & 1.63 & 6.55 & 0.129 & 18.761 & 6.434 \\
URetinex-NET~\cite{ref18}  & 0.84 & 20.69 & \second{0.12} & \best{3.99} & 35.89 & 2.16 & 4.22 & 0.340 & 79.141 & 4.726 \\
DarkIR~\cite{ref63}        & \second{0.85} & 20.81 & 0.14 & 4.41 & 45.33 & 2.43 & 4.00 & 3.322 & 14.311 & 17.366 \\
CIDNet~\cite{ref64}        & \second{0.85} & \second{20.87} & 0.13 & 4.50 & 42.04 & 2.43 & 4.28 & 1.976 & 16.412 & 23.303 \\
LoLi-IEA~\cite{ref68}      & 0.55 & 17.66 & 0.34 & 7.34 & 49.22 & 2.08 & 9.18 & 26.593 & 58.360 & 4.347 \\
LiteIE~\cite{ref4}         & 0.56 & 17.91 & 0.34 & 8.19 & \second{2.43} & 2.42 & 19.75 & \best{0.000058} & 0.0259 & \second{1.214} \\
LIME~\cite{ref5}           & 0.49 & 16.67 & 0.41 & 7.70 & 103.92 & 1.96 & \second{22.78} & -- & -- & 858.343 \\
ZeroDCE++~\cite{ref6}      & 0.59 & 18.38 & 0.31 & 8.05 & 21.46 & 2.07 & 20.15 & 0.011 & \second{0.024} & 1.806 \\
ZeroDCE~\cite{ref7}        & 0.58 & 17.99 & 0.31 & 8.05 & 24.75 & 1.94 & 20.18 & 0.079 & 10.384 & 1.604 \\
SCI~\cite{ref8}            & 0.55 & 17.25 & 0.32 & 8.26 & 6.58 & 1.96 & 21.39 & \second{0.00026} & 0.032 & \best{0.928} \\
RUAS~\cite{ref9}           & 0.47 & 13.97 & 0.34 & 7.69 & \best{0.18} & 1.44 & \best{23.23} & 0.00344 & 0.441 & 5.108 \\
EnlightenGAN~\cite{ref10}  & 0.74 & 18.48 & 0.28 & 5.24 & 76.79 & 1.44 & 3.04 & 8.637 & 32.783 & 4.212 \\
ChebyLighter~\cite{ref14}  & 0.77 & 18.83 & 0.18 & 5.38 & 29.42 & 2.19 & 4.52 & 0.178 & 22.957 & 19.73 \\
PairLIE~\cite{ref16}       & 0.77 & 19.88 & 0.23 & 4.38 & 66.75 & 1.93 & 4.71 & 0.342 & 44.695 & 2.589 \\
NoiSER~\cite{ref17}        & 0.68 & 16.47 & 0.44 & 4.15 & 72.92 & 2.30 & 2.88 & 0.00176 & \best{0.0226} & 1.283 \\
ZeroIG~\cite{ref19}        & 0.44 & 15.66 & 0.41 & 8.97 & 13.89 & \best{2.58} & 21.41 & 0.087 & 20.963 & 1.951 \\
IUJDE~\cite{ref65}         & 0.77 & 19.75 & 0.29 & 4.24 & 50.76 & 2.11 & 2.80 & 0.345 & 46.555 & 24.617 \\
\textbf{GLFS (Ours)}       & \best{0.87} & \best{27.25} & \best{0.10} & \second{4.11} & 32.57 & 2.38 & 6.30 & 7.804 & 66.824 & 14.471 \\
\bottomrule
\end{tabular}}
\end{table*}

\begin{figure*}[t]
    \centering
    \includegraphics[width=0.9\textwidth]{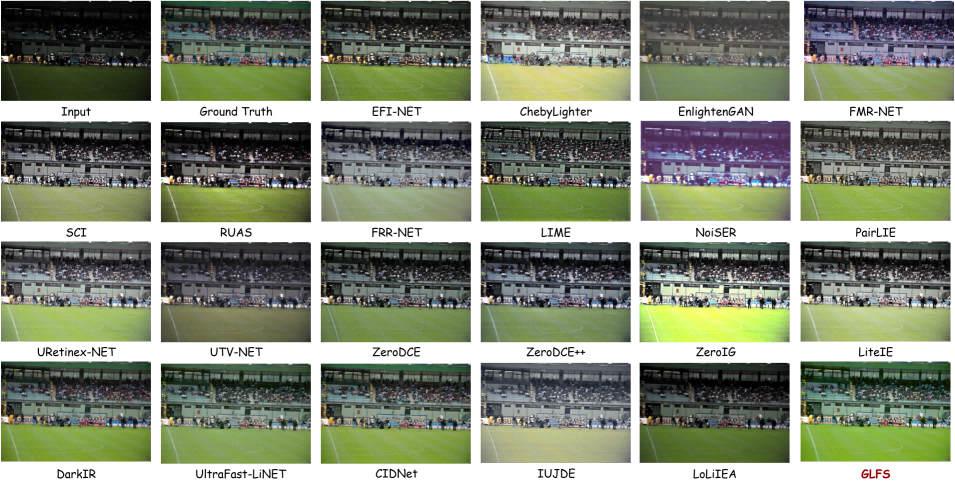}
    \caption{Visual comparison of GLFS with SOTA methods on the LOL-V2-Syn dataset.}
    \label{fig:compare_lol_v2_syn}
\end{figure*}

\begin{table*}[t]
\centering
\renewcommand{\arraystretch}{0.85}
\caption{Performance comparison on the LOL-v2-Syn dataset. The best and second-best results for each metric are highlighted in red and blue, respectively.}
\label{tab:performance_lolv2syn}
\resizebox{0.95\textwidth}{!}{
\begin{tabular}{l|cccccccccc}
\toprule
Method & SSIM$\uparrow$ & PSNR$\uparrow$ & LPIPS$\downarrow$ & NIQE$\downarrow$ & LOE$\downarrow$ & DE$\uparrow$ & EME$\uparrow$ & Params(M)$\downarrow$ & FLOPs(G)$\downarrow$ & Time(ms)$\downarrow$ \\
\midrule
FMR-NET~\cite{ref11}       & 0.88 & 20.47 & 0.13 & 4.34 & 19.54 & 0.97 & 6.22 & 0.197 & 205.658 & 8.74 \\
FRR-NET~\cite{ref12}       & 0.84 & 19.30 & 0.19 & 4.88 & 19.78 & 0.93 & 7.41 & 0.0183 & 0.665 & 5.77 \\
UTV-NET~\cite{ref13}       & 0.70 & 14.63 & 0.30 & 4.73 & 34.87 & 0.82 & 7.64 & 7.751 & 41.671 & 31.78 \\
EFI-NET~\cite{ref15}       & 0.77 & 17.28 & 0.21 & 4.23 & 44.71 & 0.82 & 7.02 & 0.129 & 18.761 & 6.434 \\
URetinex-NET~\cite{ref18}  & 0.83 & 18.76 & 0.18 & 4.26 & 23.78 & 0.93 & 10.07 & 0.340 & 79.141 & 4.726 \\
DarkIR~\cite{ref63}        & \second{0.91} & 25.22 & \second{0.09} & \second{4.21} & 21.21 & 1.01 & 5.93 & 3.322 & 14.311 & 17.366 \\
CIDNet~\cite{ref64}        & \best{0.93} & \second{25.70} & \best{0.05} & \best{4.10} & 21.57 & 1.01 & 5.95 & 1.976 & 16.412 & 23.303 \\
LoLi-IEA~\cite{ref68}      & 0.69 & 14.77 & 0.20 & 4.42 & 16.71 & 0.34 & \best{27.42} & 26.593 & 58.360 & 4.347 \\
LiteIE~\cite{ref4}         & 0.79 & 16.88 & 0.19 & 4.86 & \best{1.83} & 0.54 & 17.16 & \best{0.000058} & 0.0259 & \second{1.214} \\
LIME~\cite{ref5}           & 0.71 & 16.73 & 0.27 & 4.63 & 102.77 & 0.99 & 20.74 & -- & -- & 858.343 \\
ZeroDCE++~\cite{ref6}      & 0.81 & 17.65 & 0.19 & 4.42 & 43.49 & 0.95 & 17.50 & 0.011 & \second{0.024} & 1.806 \\
ZeroDCE~\cite{ref7}        & 0.81 & 17.71 & 0.17 & 4.37 & 42.39 & 0.77 & 17.59 & 0.079 & 10.384 & 1.604 \\
SCI~\cite{ref8}            & 0.74 & 16.69 & 0.24 & 4.67 & 31.55 & 0.73 & 18.56 & \second{0.00026} & 0.032 & \best{0.928} \\
RUAS~\cite{ref9}           & 0.63 & 13.76 & 0.30 & 5.41 & \second{2.05} & 0.29 & \second{21.18} & 0.00344 & 0.441 & 5.108 \\
EnlightenGAN~\cite{ref10}  & 0.76 & 16.61 & 0.24 & 4.24 & 94.18 & 0.36 & 3.21 & 8.637 & 32.783 & 4.212 \\
ChebyLighter~\cite{ref14}  & 0.62 & 13.42 & 0.38 & 4.24 & 176.01 & \second{1.09} & 9.64 & 0.178 & 22.957 & 19.73 \\
PairLIE~\cite{ref16}       & 0.79 & 19.07 & 0.23 & 5.19 & 65.60 & 0.86 & 5.49 & 0.342 & 44.695 & 2.589 \\
NoiSER~\cite{ref17}        & 0.64 & 14.98 & 0.48 & 5.24 & 39.63 & 0.20 & 2.30 & 0.00176 & \best{0.0226} & 1.283 \\
ZeroIG~\cite{ref19}        & 0.72 & 14.83 & 0.27 & 6.41 & 26.92 & 0.13 & 20.05 & 0.087 & 20.963 & 1.951 \\
IUJDE~\cite{ref65}         & 0.81 & 18.36 & 0.30 & 4.60 & 39.31 & 0.94 & 4.11 & 0.345 & 46.555 & 24.617 \\
\textbf{GLFS (Ours)}       & \best{0.93} & \best{26.46} & 0.17 & 4.40 & 20.75 & \best{1.22} & 11.25 & 7.804 & 66.824 & 14.471 \\
\bottomrule
\end{tabular}}
\end{table*}

\begin{figure*}[t]
    \centering
    \includegraphics[width=0.9\textwidth]{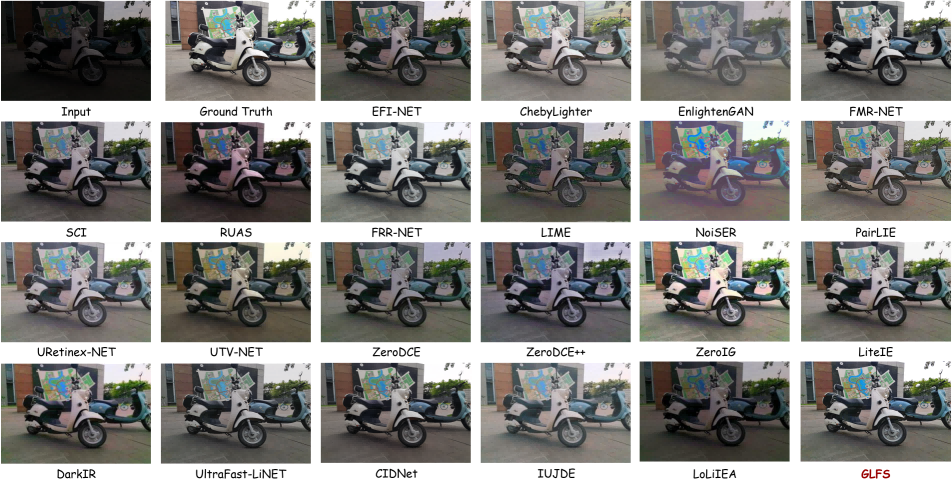}
    \caption{Visual comparison of GLFS with SOTA methods on the LSRW-Huawei dataset.}
    \label{fig:compare_lsrw_huawei}
\end{figure*}

\begin{table*}[t]
\centering
\renewcommand{\arraystretch}{0.85}
\caption{Performance comparison on the LSRW-Huawei dataset. The best and second-best results for each metric are highlighted in red and blue, respectively.}
\label{tab:performance_lsrw-huawei}
\resizebox{0.95\textwidth}{!}{
\begin{tabular}{l|cccccccccc}
\toprule
Method & SSIM$\uparrow$ & PSNR$\uparrow$ & LPIPS$\downarrow$ & NIQE$\downarrow$ & LOE$\downarrow$ & DE$\uparrow$ & EME$\uparrow$ & Params(M)$\downarrow$ & FLOPs(G)$\downarrow$ & Time(ms)$\downarrow$ \\
\midrule
FMR-NET~\cite{ref11}       & \second{0.60} & 19.69 & 0.47 & 5.19 & 30.29 & \second{2.33} & 3.05 & 0.197 & 205.658 & 8.74 \\
FRR-NET~\cite{ref12}       & 0.59 & 19.66 & 0.44 & 4.34 & 24.16 & 2.29 & 2.92 & 0.0183 & 0.665 & 5.77 \\
UTV-NET~\cite{ref13}       & 0.56 & 18.24 & 0.31 & 3.84 & 23.52 & 2.21 & 4.15 & 7.751 & 41.671 & 31.78 \\
EFI-NET~\cite{ref15}       & 0.49 & 14.91 & 0.36 & 4.22 & 20.26 & 1.60 & 4.45 & 0.129 & 18.761 & 6.434 \\
URetinex-NET~\cite{ref18}  & 0.55 & 19.53 & 0.36 & 3.68 & 27.43 & 2.18 & 4.55 & 0.340 & 79.141 & 4.726 \\
DarkIR~\cite{ref63}        & 0.54 & 18.52 & 0.36 & 3.94 & 10.26 & 2.11 & 4.83 & 3.322 & 14.311 & 17.366 \\
CIDNet~\cite{ref64}        & \second{0.60} & \second{20.81} & \second{0.30} & 4.51 & 27.39 & 2.23 & 2.55 & 1.976 & 16.412 & 23.303 \\
LoLi-IEA~\cite{ref68}      & 0.49 & 15.97 & 0.39 & 3.73 & 22.73 & 1.59 & 8.88 & 26.593 & 58.360 & 4.347 \\
LiteIE~\cite{ref4}         & 0.47 & 18.92 & 0.36 & 3.58 & \second{3.33} & 1.36 & 18.43 & \best{0.000058} & 0.0259 & \second{1.214} \\
LIME~\cite{ref5}           & 0.34 & 14.94 & 0.49 & 3.96 & 116.50 & 1.75 & \best{22.13} & -- & -- & 858.343 \\
ZeroDCE++~\cite{ref6}      & 0.47 & 16.54 & 0.38 & 3.70 & 23.72 & 1.71 & 18.56 & 0.011 & \second{0.024} & 1.806 \\
ZeroDCE~\cite{ref7}        & 0.46 & 16.34 & 0.37 & 3.62 & 22.03 & 1.54 & 18.50 & 0.079 & 10.384 & 1.604 \\
SCI~\cite{ref8}            & 0.42 & 15.09 & 0.38 & 3.94 & 6.13 & 1.33 & 18.81 & \second{0.00026} & 0.032 & \best{0.928} \\
RUAS~\cite{ref9}           & 0.34 & 12.53 & 0.44 & 4.22 & \best{0.09} & 0.82 & 18.44 & 0.00344 & 0.441 & 5.108 \\
EnlightenGAN~\cite{ref10}  & 0.53 & 18.04 & 0.36 & 3.71 & 58.77 & 1.77 & 2.36 & 8.637 & 32.783 & 4.212 \\
ChebyLighter~\cite{ref14}  & 0.54 & 17.80 & 0.38 & \second{3.18} & 46.43 & 2.17 & 3.89 & 0.178 & 22.957 & 19.73 \\
PairLIE~\cite{ref16}       & 0.56 & 19.07 & 0.38 & 3.60 & 39.99 & 1.82 & 2.81 & 0.342 & 44.695 & 2.589 \\
NoiSER~\cite{ref17}        & 0.53 & 15.77 & 0.56 & 4.51 & 38.09 & 1.92 & 1.50 & 0.00176 & \best{0.0226} & 1.283 \\
ZeroIG~\cite{ref19}        & 0.41 & 16.85 & 0.41 & 3.92 & 8.75 & 1.90 & \second{19.39} & 0.087 & 20.963 & 1.951 \\
IUJDE~\cite{ref65}         & 0.50 & 14.28 & 0.61 & 5.95 & 79.63 & 0.61 & 0.70 & 0.345 & 46.555 & 24.617 \\
\textbf{GLFS (Ours)}       & \best{0.61} & \best{21.47} & \best{0.21} & \best{3.02} & 9.40 & \best{2.48} & 11.07 & 7.804 & 66.824 & 14.471 \\
\bottomrule
\end{tabular}}
\end{table*}

\begin{figure*}[t]
    \centering
    \includegraphics[width=0.9\textwidth]{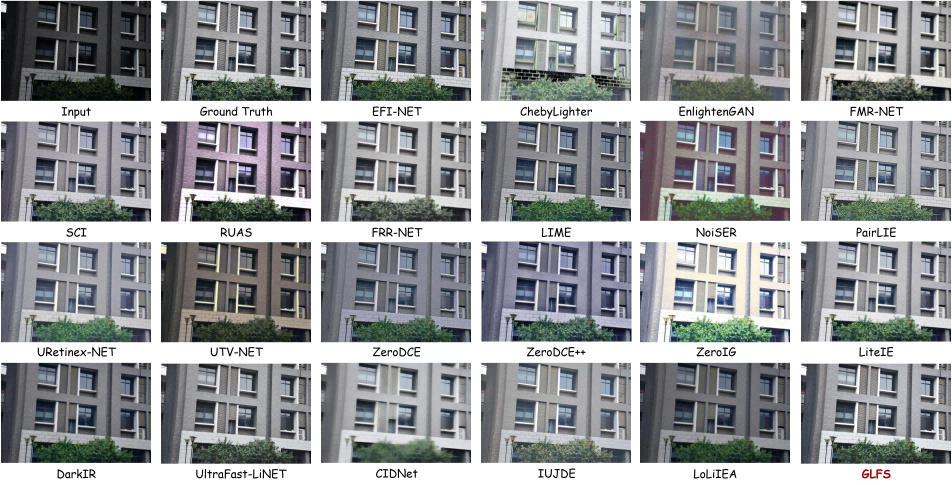}
    \caption{Visual comparison of GLFS with SOTA methods on the LSRW-Nikon dataset.}
    \label{fig:compare_lsrw_nikon}
\end{figure*}

\begin{table*}[t]
\centering
\renewcommand{\arraystretch}{0.85}
\caption{Performance comparison on the LSRW-Nikon dataset. The best and second-best results for each metric are highlighted in red and blue, respectively.}
\label{tab:performance_lsrw-nikon}
\resizebox{0.95\textwidth}{!}{
\begin{tabular}{l|cccccccccc}
\toprule
Method & SSIM$\uparrow$ & PSNR$\uparrow$ & LPIPS$\downarrow$ & NIQE$\downarrow$ & LOE$\downarrow$ & DE$\uparrow$ & EME$\uparrow$ & Params(M)$\downarrow$ & FLOPs(G)$\downarrow$ & Time(ms)$\downarrow$ \\
\midrule
FMR-NET~\cite{ref11}       & 0.50 & 17.27 & 0.49 & 5.68 & 71.24 & 1.19 & 3.40 & 0.197 & 205.658 & 8.74 \\
FRR-NET~\cite{ref12}       & \second{0.51} & \second{17.42} & 0.35 & 4.33 & 61.18 & 1.16 & 3.02 & 0.0183 & 0.665 & 5.77 \\
UTV-NET~\cite{ref13}       & 0.43 & 13.67 & 0.26 & 4.29 & 28.90 & 1.01 & 6.75 & 7.751 & 41.671 & 31.78 \\
EFI-NET~\cite{ref15}       & 0.43 & 15.79 & 0.24 & 4.15 & 36.09 & 1.05 & 5.78 & 0.129 & 18.761 & 6.434 \\
URetinex-NET~\cite{ref18}  & 0.45 & 16.64 & \second{0.22} & 3.94 & 25.87 & 0.93 & 4.52 & 0.340 & 79.141 & 4.726 \\
DarkIR~\cite{ref63}        & 0.44 & 15.52 & 0.23 & 4.02 & 13.48 & \second{1.25} & 5.75 & 3.322 & 14.311 & 17.366 \\
CIDNet~\cite{ref64}        & 0.48 & 17.03 & 0.27 & 4.73 & 46.76 & 1.14 & 3.63 & 1.976 & 16.412 & 23.303 \\
LoLi-IEA~\cite{ref68}      & 0.41 & 13.88 & 0.29 & 4.07 & 22.01 & 0.88 & \best{16.36} & 26.593 & 58.360 & 4.347 \\
LiteIE~\cite{ref4}         & 0.44 & 16.94 & \best{0.21} & 3.89 & \second{0.10} & 0.74 & 10.72 & \best{0.000058} & 0.0259 & \second{1.214} \\
LIME~\cite{ref5}           & 0.33 & 14.49 & 0.31 & 3.60 & 87.06 & 1.24 & 13.43 & -- & -- & 858.343 \\
ZeroDCE++~\cite{ref6}      & 0.43 & 15.58 & 0.23 & 3.83 & 43.97 & 1.22 & 10.82 & 0.011 & \second{0.024} & 1.806 \\
ZeroDCE~\cite{ref7}        & 0.41 & 15.02 & 0.24 & 3.71 & 51.62 & 1.02 & 10.86 & 0.079 & 10.384 & 1.604 \\
SCI~\cite{ref8}            & 0.38 & 15.27 & 0.24 & 3.90 & 25.97 & 1.04 & 12.12 & \second{0.00026} & 0.032 & \best{0.928} \\
RUAS~\cite{ref9}           & 0.37 & 13.76 & 0.29 & 4.31 & \best{0.01} & 0.23 & \second{13.44} & 0.00344 & 0.441 & 5.108 \\
EnlightenGAN~\cite{ref10}  & 0.43 & 15.32 & 0.27 & 3.78 & 89.38 & 0.71 & 2.98 & 8.637 & 32.783 & 4.212 \\
ChebyLighter~\cite{ref14}  & 0.39 & 12.97 & 0.32 & \second{3.55} & 87.05 & 1.24 & 5.66 & 0.178 & 22.957 & 19.73 \\
PairLIE~\cite{ref16}       & 0.43 & 15.52 & 0.26 & 3.96 & 55.87 & 1.05 & 4.36 & 0.342 & 44.695 & 2.589 \\
NoiSER~\cite{ref17}        & 0.44 & 15.55 & 0.52 & 4.94 & 42.85 & 0.47 & 1.93 & 0.00176 & \best{0.0226} & 1.283 \\
ZeroIG~\cite{ref19}        & 0.34 & 12.58 & 0.33 & 4.53 & 29.47 & 0.70 & 12.11 & 0.087 & 20.963 & 1.951 \\
IUJDE~\cite{ref65}         & 0.47 & 16.55 & 0.34 & 4.34 & 27.12 & 1.07 & 2.39 & 0.345 & 46.555 & 24.617 \\
\textbf{GLFS (Ours)}       & \best{0.54} & \best{17.82} & 0.23 & \best{3.11} & 10.50 & \best{1.28} & 9.80 & 7.804 & 66.824 & 14.471 \\
\bottomrule
\end{tabular}}
\end{table*}

\begin{figure*}[t]
    \centering
    \includegraphics[width=0.9\textwidth]{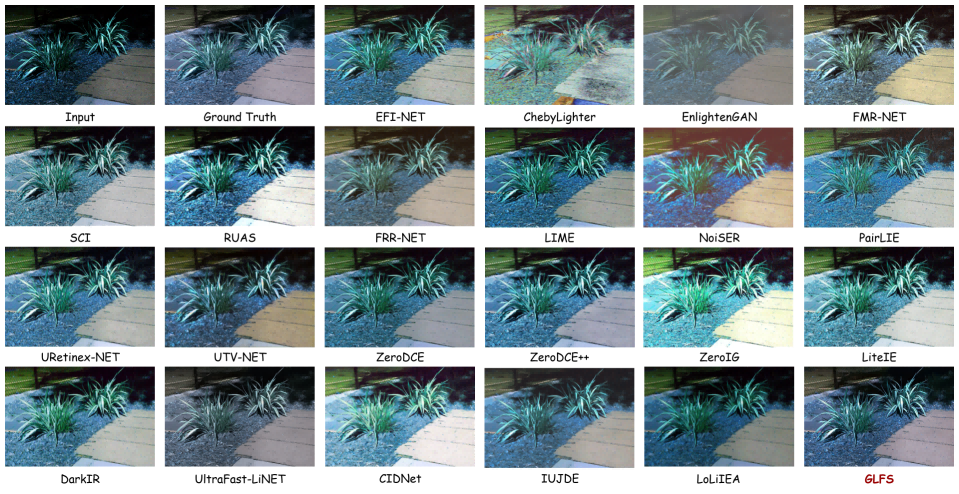}
    \caption{Visual comparison of GLFS with SOTA methods on the SID dataset.}
    \label{fig:compare_sid}
\end{figure*}

\begin{table*}[t]
\centering
\renewcommand{\arraystretch}{0.85}
\caption{Performance comparison on the SID dataset. The best and second-best results for each metric are highlighted in red and blue, respectively.}
\label{tab:performance_sid}
\resizebox{0.95\textwidth}{!}{
\begin{tabular}{l|cccccccccc}
\toprule
Method & SSIM$\uparrow$ & PSNR$\uparrow$ & LPIPS$\downarrow$ & NIQE$\downarrow$ & LOE$\downarrow$ & DE$\uparrow$ & EME$\uparrow$ & Params(M)$\downarrow$ & FLOPs(G)$\downarrow$ & Time(ms)$\downarrow$ \\
\midrule
FMR-NET~\cite{ref11}       & 0.51 & 16.78 & 0.71 & 7.27 & 120.83 & 3.75 & 3.56 & 0.197 & 205.658 & 8.74 \\
FRR-NET~\cite{ref12}       & 0.42 & 17.42 & 0.71 & 6.02 & 104.52 & 3.79 & 4.32 & 0.0183 & 0.665 & 5.77 \\
UTV-NET~\cite{ref13}       & 0.47 & 19.15 & 0.59 & 4.91 & 39.85 & \second{4.77} & 5.68 & 7.751 & 41.671 & 31.78 \\
EFI-NET~\cite{ref15}       & 0.44 & 14.06 & 0.66 & 6.19 & 63.62 & 2.11 & 3.65 & 0.129 & 18.761 & 6.434 \\
URetinex-NET~\cite{ref18}  & 0.26 & 14.12 & 0.62 & 5.43 & 51.11 & 3.70 & 29.47 & 0.340 & 79.141 & 4.726 \\
DarkIR~\cite{ref63}        & 0.26 & 13.79 & 0.74 & 6.47 & 103.89 & 3.39 & 7.88 & 3.322 & 14.311 & 17.366 \\
CIDNet~\cite{ref64}        & \second{0.55} & \second{20.86} & \second{0.49} & \second{4.50} & 138.15 & \best{5.26} & 2.52 & 1.976 & 16.412 & 23.303 \\
LoLi-IEA~\cite{ref68}      & 0.25 & 13.03 & 0.78 & 8.99 & 339.47 & 2.38 & 16.06 & 26.593 & 58.360 & 4.347 \\
LiteIE~\cite{ref4}         & 0.17 & 12.14 & 0.74 & 7.15 & 0.55 & 1.85 & 112.33 & \best{0.000058} & 0.0259 & \second{1.214} \\
LIME~\cite{ref5}           & 0.07 & 12.24 & 0.87 & 8.48 & 48.06 & 2.47 & \best{122.84} & -- & -- & 858.343 \\
ZeroDCE++~\cite{ref6}      & 0.09 & 11.96 & 0.72 & 7.28 & 0.30 & 1.77 & 110.54 & 0.011 & \second{0.024} & 1.806 \\
ZeroDCE~\cite{ref7}        & 0.09 & 11.92 & 0.72 & 7.30 & 0.41 & 1.71 & 109.66 & 0.079 & 10.384 & 1.604 \\
SCI~\cite{ref8}            & 0.09 & 11.65 & 0.71 & 7.72 & \second{0.06} & 1.33 & 105.27 & \second{0.00026} & 0.032 & \best{0.928} \\
RUAS~\cite{ref9}           & 0.08 & 11.47 & 0.73 & 8.27 & \best{0.04} & 0.90 & 102.71 & 0.00344 & 0.441 & 5.108 \\
EnlightenGAN~\cite{ref10}  & 0.43 & 14.06 & 0.74 & 8.70 & 116.62 & 1.92 & 1.06 & 8.637 & 32.783 & 4.212 \\
ChebyLighter~\cite{ref14}  & 0.31 & 14.04 & 0.61 & 6.28 & 28.42 & 3.08 & 7.10 & 0.178 & 22.957 & 19.73 \\
PairLIE~\cite{ref16}       & 0.46 & 14.84 & 0.71 & 8.51 & 63.51 & 1.76 & 2.82 & 0.342 & 44.695 & 2.589 \\
NoiSER~\cite{ref17}        & 0.18 & 11.81 & 0.80 & 13.59 & 684.92 & 1.27 & 82.61 & 0.00176 & \best{0.0226} & 1.283 \\
ZeroIG~\cite{ref19}        & 0.09 & 12.15 & 0.77 & 7.33 & 0.36 & 1.92 & \second{113.68} & 0.087 & 20.963 & 1.951 \\
IUJDE~\cite{ref65}         & 0.38 & 16.33 & 0.73 & 5.23 & 84.96 & 3.89 & 3.12 & 0.345 & 46.555 & 24.617 \\
\textbf{GLFS (Ours)}       & \best{0.56} & \best{22.46} & \best{0.41} & \best{4.20} & 16.79 & 2.03 & 17.30 & 7.804 & 66.824 & 14.471 \\
\bottomrule
\end{tabular}}
\end{table*}

The training objective of GLFS combines paired reconstruction supervision with eight categories of unsupervised constraints, forming a hybrid objective with nine main loss terms. Paired supervision provides absolute exposure and color references for the enhancement mapping, while adversarial and cycle-consistency constraints regularize the cross-domain distributions. In addition, exposure, color, edge, and physical smoothness terms impose reference-free constraints on the spatial continuity of physical quantities as well as color and structural fidelity, thereby continuing to provide effective gradients in regions where reference signals are unavailable or imperfectly aligned. To avoid the burden of tuning weights for excessively fine-grained sub-losses, we group sub-terms with the same physical meaning according to their semantics and ultimately optimize the bidirectional generative adversarial network using nine main losses. Let $I_n$ and $I_l$ denote the normal-light and low-light images, respectively. $G_A$ denotes the darkening generator that maps normal-light images to the low-light domain, whereas $G_B$ denotes the core enhancement generator that maps low-light images to the normal-light domain. $D_A$ and $D_B$ denote the corresponding multi-scale discriminators. The enhanced and darkened outputs are defined as $\hat{I}_e=G_B(I_l)$ and $\hat{I}_d=G_A(I_n)$, respectively.

1) \textbf{Paired Reconstruction Supervision.}
When paired low-light and normal-light images are pixel-wise aligned, we directly supervise the enhancement direction at the pixel, structural, and perceptual levels, while imposing a down-weighted pixel-level constraint on the darkening direction:

\begin{equation}
\begin{aligned}
\mathcal{L}_{\mathrm{pair}}
=
& \lambda_{l_1}\left\| \hat{I}_e-I_n \right\|_1
+
\lambda_s\left(1-\mathrm{SSIM}(\hat{I}_e,I_n)\right) \\
& +
\lambda_{\mathrm{vgg}}
\sum_{l=1}^{L}
\left\|
\phi_l(\hat{I}_e)-\phi_l(I_n)
\right\|_1
+
\varsigma
\left\| \hat{I}_d-I_l \right\|_1 ,
\end{aligned}
\end{equation}
where $\phi_l$ denotes the activation features extracted from multiple layers of VGG-19, and $\varsigma$ is the down-weighting coefficient for the darkening direction. This term serves as a key complement to a purely unpaired learning paradigm by directly anchoring the absolute brightness and color references, thereby fundamentally avoiding the systematic biases introduced by the gray-world assumption and fixed target exposure in scenes with severe color casts.

2) \textbf{Unsupervised Constraints.}
To bring the distributions of the two generated domains closer without relying on paired data, we adopt a least-squares adversarial loss and combine the multi-scale PatchGAN real/fake constraints in both translation directions into a unified adversarial objective:
\begin{equation}
\mathcal{L}_{\mathrm{adv}}
=
\mathbb{E}\left[\left(D_B\left(G_B(I_l)\right)-1\right)^2\right]
+
\mathbb{E}\left[\left(D_A\left(G_A(I_n)\right)-1\right)^2\right].
\end{equation}

To prevent mode collapse and cross-domain structural misalignment under purely adversarial training, we further impose both cycle-consistency and identity constraints. The former requires the two translation paths, darkening$\rightarrow$enhancement and enhancement$\rightarrow$darkening, to reconstruct their original inputs, whereas the latter requires each generator to preserve an input that already belongs to its target domain. Since both constraints are measured using the pixel-wise $\ell_1$ distance and jointly maintain cross-domain semantic consistency, we combine them into a unified cross-domain consistency loss:
\begin{equation}
\begin{aligned}
\mathcal{L}_{\mathrm{dc}}
={}&
\left\|G_B\left(G_A(I_n)\right)-I_n\right\|_1
+
\left\|G_A\left(G_B(I_l)\right)-I_l\right\|_1
\\
&+
\eta\left(
\left\|G_A(I_l)-I_l\right\|_1
+
\left\|G_B(I_n)-I_n\right\|_1
\right),
\end{aligned}
\end{equation}
where $\eta$ denotes the relative weight of the identity term.

To further constrain the cycle reconstructions to preserve high-level structural and texture consistency at the semantic level, we introduce a multi-layer perceptual loss based on VGG-19. To control memory consumption and computational overhead, we use the activation features from the first $M$ layers of VGG-19, denoted as $\{\phi_m\}_{m=1}^{M}$, with $M=3$ in our experiments. The $\ell_1$ distance is computed at each layer and accumulated over both translation directions:
\begin{equation}
\begin{aligned}
\mathcal{L}_{\mathrm{per}}
=
\sum_{m=1}^{M}
\Big(
&
\left\|
\phi_m\left(G_B\left(G_A(I_n)\right)\right)
-
\phi_m(I_n)
\right\|_1
\\
&+
\left\|
\phi_m\left(G_A\left(G_B(I_l)\right)\right)
-
\phi_m(I_l)
\right\|_1
\Big).
\end{aligned}
\end{equation}

Since pixel-level cycle-consistency losses tend to produce over-smoothed results in high-frequency regions, we further impose a structural similarity loss on the cycle reconstructions. Specifically, we use window-based structural similarity as the perceptual optimization objective:
\begin{equation}
\begin{aligned}
\mathcal{L}_{\mathrm{ssim}}
={}&
\frac{1}{2}
\left[
1-
\mathrm{SSIM}\left(
G_B\left(G_A(I_n)\right),
I_n
\right)
\right]
\\
&+
\frac{1}{2}
\left[
1-
\mathrm{SSIM}\left(
G_A\left(G_B(I_l)\right),
I_l
\right)
\right].
\end{aligned}
\end{equation}

To address the issue of local exposure imbalance highlighted in the abstract, we introduce a unified exposure and spatial consistency objective. In the enhancement domain, the squared deviation between the local mean over $p\times p$ regions ($p=16$) and the target exposure $E_n=0.55$ is penalized to encourage the entire image to uniformly reach a natural brightness level. In the darkening domain, $E_l=0.20$ is adopted as the target exposure to maintain separation between the brightness distributions of the two domains. In addition, four-directional difference consistency is measured after $4\times$ pooling to prevent abrupt spatial variations introduced by enhancement:

\begin{equation}
\begin{aligned}
\mathcal{L}_{\mathrm{exp}}
={}&
\left\|
M_p(I_n)-E_n
\right\|_2^2
+
\zeta_d
\left\|
M_p(I_l)-E_l
\right\|_2^2
\\
&+
\zeta_s
\sum_r
\left\|
D_r\left(M_4(I_n)\right)
-
D_r\left(M_4(I_l)\right)
\right\|_2^2 .
\end{aligned}
\end{equation}
where $M_p(\cdot)$ denotes $p\times p$ average pooling performed independently on each channel in the $[0,1]$ range, and $D_r$, with $r\in\{\leftarrow,\rightarrow,\uparrow,\downarrow\}$, denotes the four directional difference operators. $\zeta_d$ and $\zeta_s$ are the relative weights of the corresponding sub-terms. It should be noted that the channel-wise exposure constraint may weakly align the global channel means in a statistical sense. However, this objective operates only on coarse-grained local means over $16\times16$ regions, whereas the subsequent angular constraint strictly preserves hue at the pixel level. The two objectives therefore operate at complementary spatial frequencies and levels of constraint, preventing the exposure constraint from enforcing pixel-level hue neutralization as in the gray-world assumption.

Furthermore, to overcome the color distortions caused by the conventional gray-world assumption or simple channel-wise constraints, we propose a color vector angular loss $\mathcal{L}_{\mathrm{col}}$. In low-light enhancement, increasing pixel brightness essentially corresponds to increasing the magnitude of the RGB vector. Directly constraining the absolute RGB-value difference between the input and enhanced images may restrict the desired brightness amplification, whereas the gray-world assumption may undesirably shift scenes with inherent warm or cool tones toward gray. Therefore, we treat the RGB triplet of each pixel as a three-dimensional vector and constrain its direction to preserve hue while allowing its magnitude to vary with brightness. Specifically, we compute the cosine similarity between the RGB vectors of the input low-light image $I_l$ and the enhanced output $\hat{I}_e$ at each pixel $x$ as

\begin{equation}
\mathcal{L}_{\mathrm{col}}
=
1-
\frac{1}{|\Omega|}
\sum_{x\in\Omega}
\frac{
\hat{I}_e(x)I_{l}(x)
}{
\max\left(
\left\|\hat{I}_e(x)\right\|_2
\left\|I_{l}(x)\right\|_2,
\epsilon
\right)
},
\label{eq:color_angular_loss}
\end{equation}
where $\epsilon=10^{-7}$ is a small constant introduced to avoid division by zero. This loss encourages the angle between the RGB vectors before and after enhancement to approach zero, corresponding to a cosine similarity close to one. Consequently, while the spatial gain field can substantially increase image visibility, the intrinsic hue of each pixel, represented by the vector direction, remains anchored. This formulation effectively suppresses color shifts during enhancement without relying on the absolute color values of reference images. In implementation, the final color objective combines the proposed color vector angular loss with an image-level channel-mean balancing regularizer with a weight of $0.1$. The latter suppresses channel drift only at the global image-mean level and makes a much smaller contribution than the angular term, without imposing any pixel-level hue constraint.

To preserve local high-frequency details under reference-free or imperfectly aligned conditions, we design a luminance edge loss $\mathcal{L}_{\mathrm{edge}}$. Conventional image gradient losses, such as total variation (TV) loss or simple adjacent-pixel differences, are susceptible to color variations and amplified noise in extremely dark regions. To mitigate these effects, we first convert the RGB image to the luminance Y channel, defined as $\mathrm{Y}(I)=0.299R+0.587G+0.114B$, thereby excluding the interference of color variations from structural evaluation. We then employ $3\times3$ Sobel operators, $S_x$ and $S_y$, to extract the gradient magnitude, providing a more robust structural response than simple pixel differences:

\begin{equation}
\nabla \mathrm{Y}(I)
=
\sqrt{
\left(S_x * \mathrm{Y}(I)\right)^2
+
\left(S_y * \mathrm{Y}(I)\right)^2
}
+\epsilon .
\label{eq:luminance_gradient}
\end{equation}

The resulting edge loss is formulated in a reference-free manner. Specifically, the gradient magnitude of the Y channel of the enhanced image is directly matched to that of the input low-light image, whose gradient is detached and treated as a constant target. The same constraint is further imposed on the two cycle-reconstruction paths with a weight of $1/2$:

\begin{equation}
\begin{aligned}
\mathcal{L}_{\mathrm{edge}}
={}&
\left\|
\nabla \mathrm{Y}\!\left(G_B(I_l)\right)
-
\nabla \mathrm{Y}(I_l)
\right\|_1
\\
&+
\frac{1}{2}
\sum_{(u,u^\ast)}
\left\|
\nabla \mathrm{Y}(u)
-
\nabla \mathrm{Y}(u^\ast)
\right\|_1 ,
\end{aligned}
\label{eq:edge_loss}
\end{equation}
where $(u,u^\ast)$ denotes the edge-matching pairs in the cycle-reconstruction paths, i.e.,
$\left(G_B(G_A(I_n)),I_n\right)$ and
$\left(G_A(G_B(I_l)),I_l\right)$.
Using the gradients of the low-light input itself rather than those of a reference image allows this term to remain applicable in a fully reference-free setting and makes it complementary to paired supervision. In implementation, the final edge objective additionally incorporates a one-sided gradient-preservation regularizer with a weight of $0.2$, given by
$\max\!\left(0,\left|\nabla I_l\right|-\left|\nabla I_n\right|\right)$.
This term penalizes only the loss of high-frequency structures without penalizing structural enhancement. Intuitively, the Sobel operator is more responsive to diagonal edges and complex structures, while computing gradients in the Y channel effectively avoids spurious chromatic edges caused by brightness enhancement. The bidirectional $\ell_1$ penalty not only suppresses the loss of existing high-frequency structures, such as architectural contours and textures, but also penalizes false edges introduced by abrupt gain variations or noise amplification, thereby substantially improving local structural fidelity and SSIM.

For the physical outputs of the GIF module, the three originally parallel TV smoothness terms for the enhanced result, intrinsic illumination field, and Gaussian opacity are grouped according to their shared objective of enforcing the spatial continuity of physical quantities, yielding a unified physical smoothness loss:
\begin{equation}
\mathcal{L}_{\mathrm{smooth}}
=
\rho_1 \mathrm{TV}(I_e)
+
\rho_2 \mathrm{TV}(L)
+
\rho_3 \mathrm{TV}(\alpha'),
\label{eq:smooth_loss}
\end{equation}
where
$
\mathrm{TV}(Z)
=
\mathbb{E}\left[\left|\partial_x Z\right|\right]
+
\mathbb{E}\left[\left|\partial_y Z\right|\right]
$
denotes the mean differences along the horizontal and vertical directions. For the intrinsic illumination field $L$, squared differences are used instead to impose a stronger penalty on high-frequency fluctuations. $\rho_{\ast}$ denotes the weight of each sub-term. Both $L$ and $\alpha'$ are evaluated at the bottleneck resolution $H'\times W'$, whereas $I_e$ is evaluated at the full image resolution.

Combining the above nine main losses, the final training objective of GLFS is formulated as

\begin{equation}
\begin{aligned}
\mathcal{L}_{\mathrm{total}}
={}&
\lambda_{\mathrm{adv}}\mathcal{L}_{\mathrm{adv}}
+
\lambda_{\mathrm{dc}}\mathcal{L}_{\mathrm{dc}}
+
\lambda_{\mathrm{per}}\mathcal{L}_{\mathrm{per}} \\
&+
\lambda_{\mathrm{ssim}}\mathcal{L}_{\mathrm{ssim}}
+
\lambda_{\mathrm{exp}}\mathcal{L}_{\mathrm{exp}}
+
\lambda_{\mathrm{col}}\mathcal{L}_{\mathrm{col}} \\
&+
\lambda_{\mathrm{edge}}\mathcal{L}_{\mathrm{edge}}
+
\lambda_{\mathrm{smooth}}\mathcal{L}_{\mathrm{smooth}}
+
\mathcal{L}_{\mathrm{pair}} .
\label{eq:total_loss}
\end{aligned}
\end{equation}
The optimal configurations of all weighting coefficients are provided in Section~\ref{sec:experiments}. The entire model is optimized using Adam with automatic mixed-precision (AMP) training. At each iteration, the discriminators are first frozen while the generators are updated, after which the generators are frozen while the discriminators are updated. Gradient clipping with a maximum norm of $1.0$ is applied to both generator and discriminator updates to ensure stable training.

\begin{figure}[t]
    \centering
    \includegraphics[width=\columnwidth]{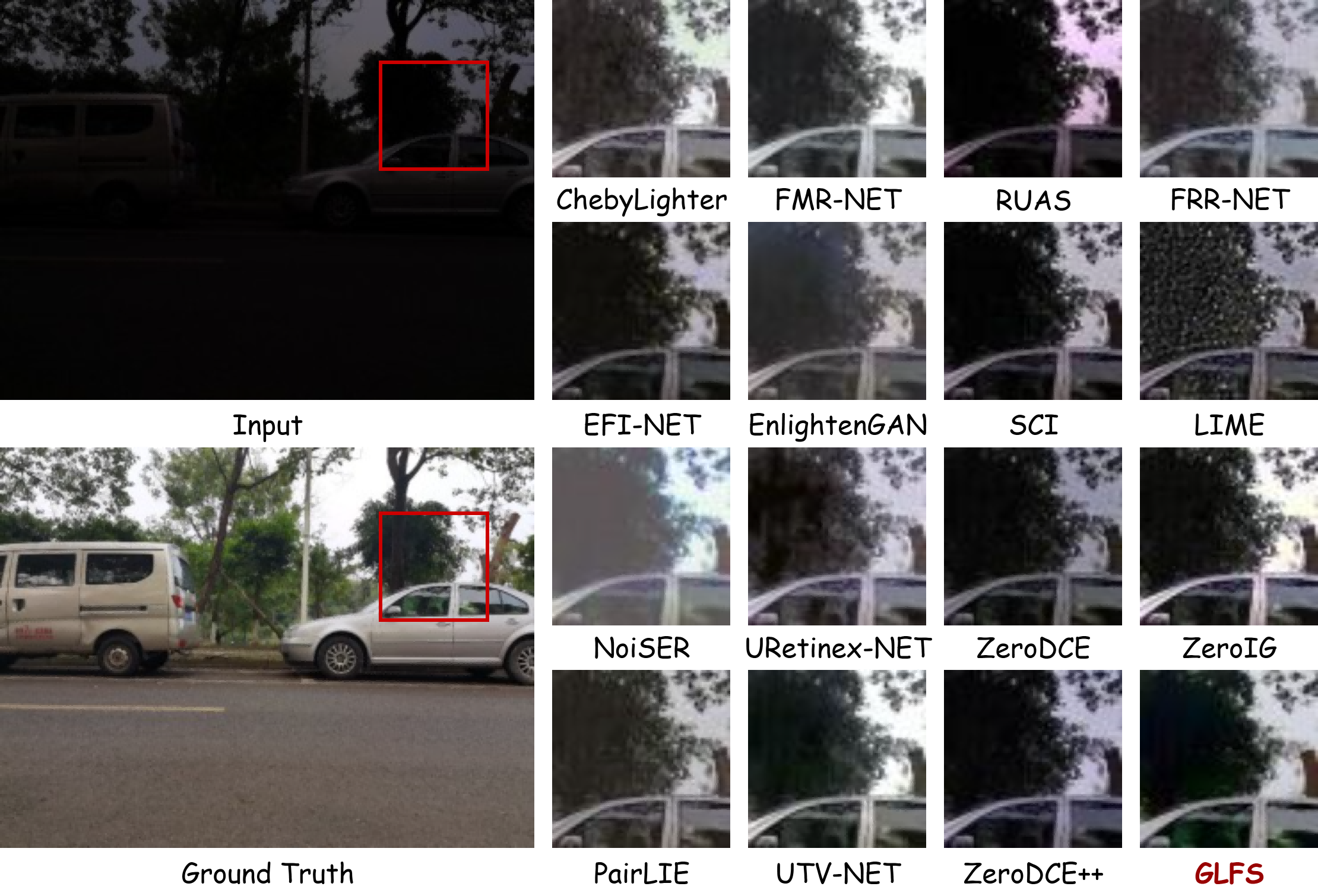}
    \caption{Detailed feature comparison of GLFS with SOTA methods on the LSRW-Huawei dataset. For each method, the enlarged view corresponds to the region marked by the red bounding box.}
    \label{fig:detail_lsrw_huawei}
\end{figure}

\section{Experiments}
\label{sec:experiments}

\subsection{Experimental Setup}
\label{sec:experimental_setup}

\noindent\textbf{Datasets.}
To systematically evaluate the effectiveness and generalization of the proposed GLFS across different degradation distributions, imaging devices, and extreme low-light conditions, we extend the evaluation to six widely used low-light image enhancement benchmarks, including LOL-v1~\cite{ref26}, LOL-v2-Real, LOL-v2-Synthetic~\cite{ref66}, LSRW-Huawei, LSRW-Nikon~\cite{ref27}, and SID~\cite{ref67}. LOL-v1 contains 500 pairs of real low-light and normal-light images, with 485 pairs for training and 15 pairs for testing. LOL-v2-Real and LOL-v2-Synthetic contain 689/100 and 900/100 training/testing pairs, respectively. LSRW is a large-scale real-world paired dataset captured using different camera devices. Its Huawei P40 Pro and Nikon D7500 subsets contain 2450/30 and 3150/20 training/testing pairs, respectively. SID targets extreme low-light imaging conditions and contains short-exposure low-light images paired with corresponding long-exposure reference images.

These six benchmarks cover distinct data distributions. LOL-v1 and LOL-v2-Real primarily represent low-light degradations in real-world scenes, whereas LOL-v2-Synthetic introduces synthetically generated illumination variations, allowing evaluation of the distribution gap between real and synthetic degradations. LSRW-Huawei and LSRW-Nikon are captured using different camera devices with distinct sensor responses, imaging pipelines, and noise characteristics, providing a direct test of robustness to cross-device domain shift. SID represents a more severe degradation regime than conventional RGB low-light datasets, with stronger underexposure and noise. Collectively, these benchmarks cover real and synthetic degradations, different imaging devices, and both conventional and extreme low-light conditions, enabling a more comprehensive evaluation of model robustness to domain shifts.

For datasets containing aligned low-light and normal-light image pairs, the original pairing relationships are preserved during training. To ensure spatial consistency for pixel-level supervision, a one-to-one correspondence between each low-light image and its reference image is established based on filenames. The same geometric augmentations, including random cropping and horizontal flipping, are then applied to both images in each pair. During testing, the original pairwise correspondence is also preserved for computing full-reference metrics, including PSNR, SSIM, and LPIPS.

\noindent\textbf{Implementation Details.}
The entire model is implemented in PyTorch, with end-to-end training and inference conducted on a single NVIDIA RTX 4090 D GPU. During data preprocessing, the original images are first resized to $286\times286$, from which $256\times256$ training patches are randomly cropped. Random horizontal flipping is applied with a probability of 0.5 for data augmentation. For the network architecture, the base channel width of the GS-ViT generator is set to 64, and the number of stacked GS-ViT blocks is set to $N=4$.

For optimization, we use Adam for both the generator and discriminator, with $\beta_1=0.5$ and $\beta_2=0.999$. The initial learning rate is set to $2\times10^{-4}$. A linear decay schedule is adopted, with the learning rate kept constant for the first 60 epochs and then linearly decayed to zero over the following 60 epochs. The model is trained for a total of 120 epochs with a batch size of 4. The adversarial objective follows the least-squares GAN formulation, and an image buffer with a capacity of 50 is maintained to stabilize discriminator training.

For the multi-objective joint loss, the loss weights are selected based on ablation studies and set to $\lambda_{\mathrm{adv}}=1.0$, $\lambda_{\mathrm{dc}}=10.0$, $\lambda_{\mathrm{per}}=5.0$, $\lambda_{\mathrm{ssim}}=3.0$, $\lambda_{\mathrm{exp}}=1.0$, $\lambda_{\mathrm{col}}=1.0$, $\lambda_{\mathrm{edge}}=1.0$, and $\lambda_{\mathrm{smooth}}=1.0$. The internal weighting coefficients of the composite losses are set to $\eta=0.5$, $\zeta_d=0.3$, $\zeta_s=1.0$, $\rho_1=0.005$, $\rho_2=1.0$, and $\rho_3=0.5$. For the paired reconstruction supervision $\mathcal{L}_{\mathrm{pair}}$, the weights of the pixel-level $\ell_1$, structural $\ell_{\mathrm{ssim}}$, and perceptual $\ell_{\mathrm{vgg}}$ terms are set to $\lambda_{l_1}=10.0$, $\lambda_s=3.0$, and $\lambda_{\mathrm{vgg}}=1.0$, respectively. The attenuation coefficient for the darkening direction is set to $\varsigma=0.5\lambda_{l_1}$. These weights balance flexible feature representation in the Transformer with the physical constraints imposed by the Gaussian light field. Paired supervision provides absolute exposure and color references, while the color and structural regularizers further reduce color shifts and structural degradation associated with unpaired training.

The evaluation is conducted on a platform equipped with an AMD EPYC 7F72 CPU (24 cores and 48 threads) and two NVIDIA RTX 4090 D GPUs, each with 24 GB of memory. The software environment consists of PyTorch 2.4.0 and CUDA 12.4. For all neural network-based methods, the inference batch size is uniformly set to 1, and all models are evaluated in FP32 precision with TF32 disabled. All image quality metrics are computed at the original image resolutions. Specifically, the image resolutions of LOL-v1, LOL-v2-Real, LOL-v2-Synthetic, LSRW-Huawei, LSRW-Nikon, and SID are $600\times400$, $600\times400$, $384\times384$, $960\times720$, $960\times640$, and $960\times512$, respectively.

\noindent\textbf{Evaluation Metrics.}
To comprehensively evaluate the enhanced images in terms of pixel-level fidelity, perceptual quality, and visual naturalness, we adopt seven widely used metrics for image restoration and enhancement, including three full-reference metrics and four no-reference metrics. For datasets with paired ground-truth (GT) images, Peak Signal-to-Noise Ratio (PSNR), Structural Similarity Index Measure (SSIM), and Learned Perceptual Image Patch Similarity (LPIPS) are used to measure the agreement between the enhanced images and their references. Specifically, PSNR measures pixel-level fidelity, while SSIM evaluates similarity in terms of luminance, contrast, and structure. LPIPS measures perceptual similarity based on the feature-space distance between the enhanced and reference images, using a VGG backbone in this study.

To evaluate image naturalness, contrast, and information content without reference images, particularly for unpaired data and real-world scenarios, we further employ four no-reference metrics: Natural Image Quality Evaluator (NIQE), Lightness Order Error (LOE), Discrete Entropy (DE), and Enhancement Measure Estimation (EME). NIQE measures image naturalness based on the deviation from natural-image statistics, while LOE evaluates the preservation of relative lightness order during enhancement. DE characterizes the information content of an image, and EME measures local contrast. Together, these metrics provide complementary assessments of reconstruction fidelity, perceptual quality, and visual naturalness across different evaluation settings.

In addition to the above seven image quality metrics, we report the number of parameters (Params), floating-point operations (FLOPs), and inference time to evaluate model complexity and computational efficiency. Params characterize the model scale, while FLOPs estimate the theoretical computational cost for a given input size. Inference time measures the actual runtime cost under the specified hardware and implementation settings, accounting for factors such as network architecture, parallelization, and operator efficiency. As GLFS prioritizes restoration quality rather than extreme model compression, Params and FLOPs are treated as efficiency indicators rather than primary optimization objectives. We therefore examine whether the performance gains introduced by explicit Gaussian light field modeling can be achieved with a reasonable computational cost.

\begin{table}[t]
\centering
\caption{Fine-grained ablation study on the physical-prior components of GLFS. The best and second-best results for each metric are highlighted in red and blue, respectively.}
\label{tab:component_ablation}
\resizebox{\columnwidth}{!}{
\begin{tabular}{lccccc}
\toprule
Model Variants & MGT & GSA & GIF & SSIM$\uparrow$ & PSNR$\uparrow$ \\
\midrule
w/o MGT (Single-scale)    & $\times$     & $\checkmark$ & $\checkmark$ & \second{0.83} & 24.07 \\
w/o GSA (Standard MHSA)   & $\checkmark$ & $\times$     & $\checkmark$ & 0.79 & 23.92 \\
w/o GIF (No Gain Field)   & $\checkmark$ & $\checkmark$ & $\times$     & 0.81 & \second{24.10} \\
\textbf{Full GLFS (Ours)} & $\checkmark$ & $\checkmark$ & $\checkmark$ & \best{0.84} & \best{24.38} \\
\bottomrule
\end{tabular}}
\end{table}

\begin{figure}[t]
\centering
\includegraphics[width=\columnwidth]{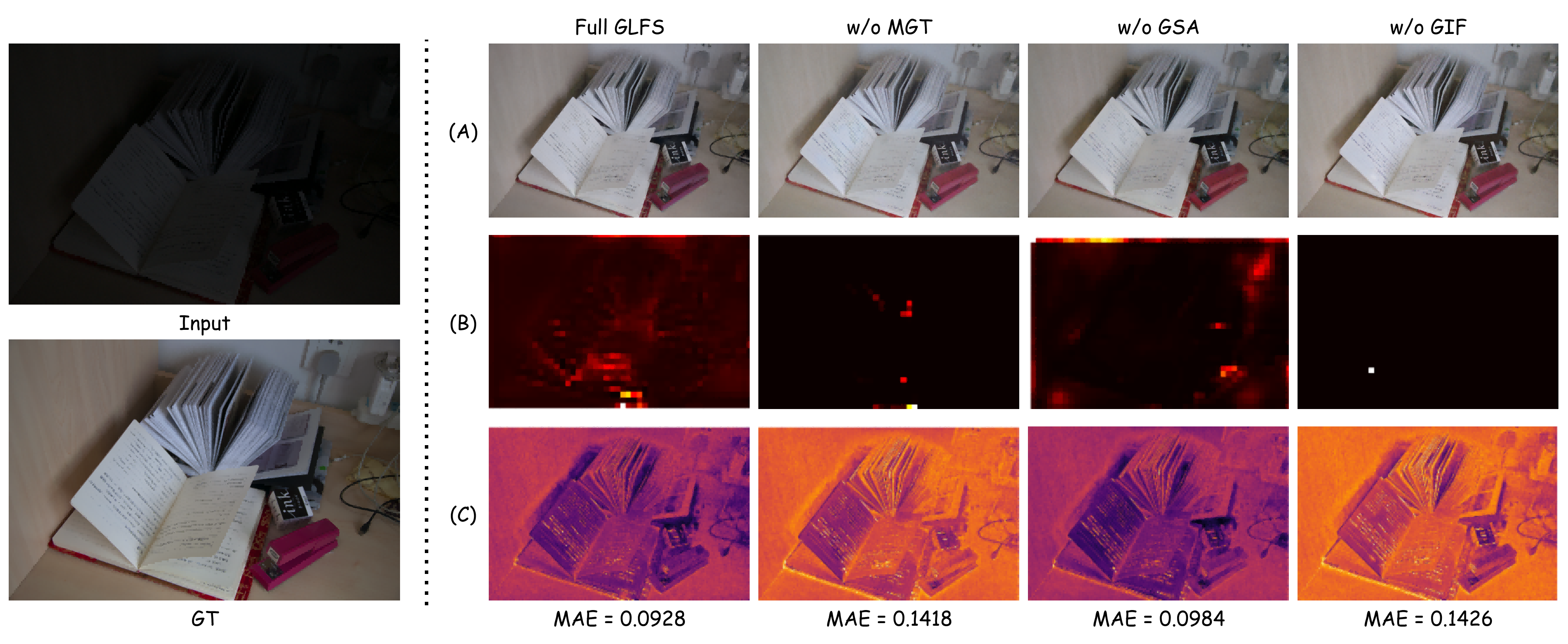}
\caption{Visual ablation comparison of the core components. (A) Enhancement results. (B) Attention maps of the center token in Block 3. (C) Absolute error maps ($|\mathrm{Output}-\mathrm{GT}|$).}
\label{fig:component_ablation}
\end{figure}

\subsection{Performance Comparison}

To comprehensively evaluate the enhancement performance of GLFS, we conduct qualitative and quantitative comparisons with 20 representative LLIE methods listed in Tables~\ref{tab:performance_lolv1}--\ref{tab:performance_sid} on the six benchmark datasets~\cite{ref4,ref5,ref6,ref7,ref8,ref9,ref10,ref11,ref12,ref13,ref14,ref15,ref16,ref17,ref18,ref19,ref20,ref63,ref64,ref65,ref68}. These methods cover various technical paradigms, including traditional optimization, zero-reference learning, unsupervised enhancement, supervised restoration, and lightweight network design. The corresponding visual comparisons are presented in Figs.~\ref{fig:compare_lol_v1}--\ref{fig:detail_lsrw_huawei}.

It should be noted that all performance metrics of the compared methods in Tables~\ref{tab:performance_lolv1}--\ref{tab:performance_sid} are recomputed by the authors following the unified evaluation protocol described in Sec.~\ref{sec:experimental_setup}, rather than directly taken from the original papers. Among the 120 evaluation settings formed by 20 methods and six datasets, 18 use retrained model weights, 15 use publicly released weights after retraining was attempted, and the remaining 87 directly use publicly released weights or the original algorithms. The final weight sources and retraining status of each method on different datasets are detailed in Table~S2-1 of Supplementary Material S2. Since GLFS is trained with paired image supervision combined with no-reference physical regularization, whereas some zero-reference or unsupervised baselines are evaluated using their publicly released weights, the training supervision conditions are not identical across all methods. In addition, some reevaluated results differ from those reported in the original papers. Due to differences in evaluation settings, these values should not be considered directly equivalent. Therefore, the comparative conclusions drawn in this work are valid only under the specified unified evaluation protocol.

As shown in Figs.~\ref{fig:compare_lol_v1}--\ref{fig:compare_lol_v2_syn}, different methods exhibit distinct enhancement behaviors on LOL-v1, LOL-v2-Real, and LOL-v2-Synthetic. Methods such as ZeroDCE, ZeroDCE++, SCI, and RUAS still show insufficient enhancement in some extremely dark regions. EnlightenGAN and NoiSER produce varying degrees of haziness, overexposure, or color shifts in certain scenes, while the strong illumination amplification of LIME can also amplify noise in dark regions. In comparison, recent methods such as DarkIR and CIDNet provide more stable overall restoration. GLFS improves the visibility of dark regions while preserving color relationships and structural edges consistent with the reference images, achieving a stable balance between exposure uniformity and detail preservation.

As shown in Figs.~\ref{fig:compare_lsrw_huawei}--\ref{fig:compare_sid}, LSRW-Huawei, LSRW-Nikon, and SID cover different imaging devices and challenging low-light conditions. Some competing methods exhibit inconsistent brightness correction across datasets, along with localized color shifts and oversmoothed textures. These artifacts suggest that differences in sensor response, noise characteristics, and image-processing pipelines can affect restoration performance. In contrast, GLFS produces more natural exposure levels and better preserves structural details in the examples shown. These improvements are particularly apparent around building edges, vehicle contours, and vegetation textures.

Figure~\ref{fig:detail_lsrw_huawei} provides a closer comparison of fine-detail restoration using magnified image crops. ChebyLighter, FMR-Net, and FRR-Net effectively restore overall brightness, but some local edges appear slightly blurred or washed out. RUAS, SCI, and ZeroDCE leave certain regions under-enhanced or exhibit localized color shifts, whereas LIME produces more noticeable noise amplification. In comparison, GLFS better preserves vehicle contours, foliage textures, and background structures without introducing noticeable localized overexposure. Overall, the qualitative results across the six datasets suggest that GLFS achieves a favorable balance between exposure correction, color fidelity, and structural preservation under diverse degradation conditions.

Tables~\ref{tab:performance_lolv1}--\ref{tab:performance_sid} summarize the quantitative results across the six datasets. Under the training and evaluation settings adopted in this study, GLFS achieves the highest PSNR on all six benchmarks, with values of 24.38, 27.25, 26.46, 21.47, 17.82, and 22.46~dB on LOL-v1, LOL-v2-Real, LOL-v2-Synthetic, LSRW-Huawei, LSRW-Nikon, and SID, respectively. It also delivers competitive results in terms of SSIM, LPIPS, NIQE, and DE on several datasets. Compared with recent methods such as DarkIR, CIDNet, and IUJDE, GLFS demonstrates strong pixel-level reconstruction fidelity on these benchmarks. However, it does not outperform all competing methods across every evaluation metric. For instance, on LOL-v2-Synthetic, GLFS yields higher LPIPS and NIQE values than CIDNet. These differences in metric rankings highlight that pixel fidelity, perceptual similarity, visual naturalness, and local contrast capture distinct aspects of enhancement quality. We therefore assess enhancement performance using both quantitative metrics and qualitative comparisons.

\begin{table}[t]
\centering
\caption{Ablation results of the core architectures. Red and blue indicate the best and second-best results for each metric, respectively.}
\label{tab:architecture_ablation}
\resizebox{\columnwidth}{!}{
\begin{tabular}{lcccccc}
\hline
Model Architecture & SSIM$\uparrow$ & PSNR$\uparrow$ & LPIPS$\downarrow$ & NIQE$\downarrow$ & Params (M)$\downarrow$ & Time (s)$\downarrow$ \\
\hline
Base-CNN &
0.79 &
23.23 &
0.27 &
4.92 &
11.37 &
\textcolor{red}{0.008} \\
Base-ViT &
\textcolor{blue}{0.82} &
\textcolor{blue}{23.46} &
\textcolor{blue}{0.24} &
\textcolor{blue}{4.58} &
\textcolor{blue}{7.97} &
0.023 \\
GLFS (Ours) &
\textcolor{red}{0.84} &
\textcolor{red}{24.38} &
\textcolor{red}{0.12} &
\textcolor{red}{4.14} &
\textcolor{red}{7.80} &
\textcolor{blue}{0.014} \\
\hline
\end{tabular}
}
\end{table}

\begin{figure}[t]
\centering
\includegraphics[width=\columnwidth]{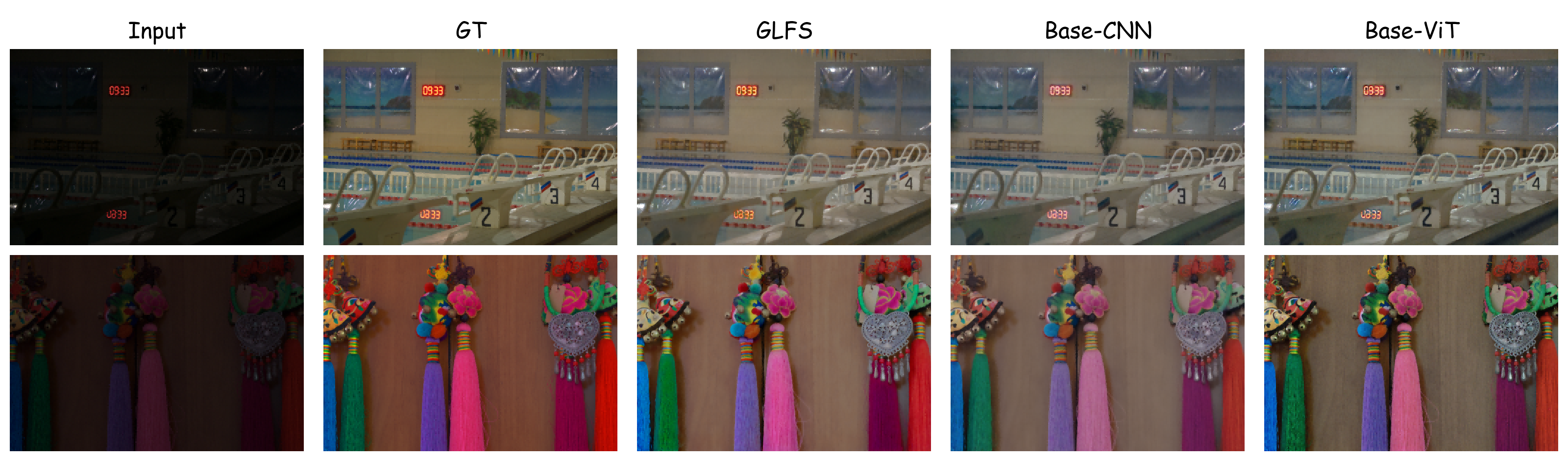}
\caption{Visual comparisons of the core architecture ablation. From left to right: low-light input, reference image, Base-CNN, Base-ViT, and GLFS.}
\label{fig:architecture_ablation}
\end{figure}

In terms of computational complexity, GLFS has 7.804~M parameters and requires 66.824~G FLOPs, with a measured inference latency of 14.471~ms. Although GLFS has more parameters and requires more computation than lightweight methods such as LiteIE and SCI, its measured inference latency is lower than that of DarkIR and CIDNet, while its computational cost remains lower than that of FMR-Net. Rather than prioritizing minimal model size and computational cost, GLFS seeks to balance enhancement quality and inference efficiency through the incorporation of an interpretable Gaussian light-field prior.

\begin{table}[t]
\centering
\caption{Ablation results of different physical-prior modeling strategies for the illumination field. Red and blue indicate the best and second-best results for each metric, respectively.}
\label{tab:physical_prior_ablation}
\resizebox{\columnwidth}{!}{
\begin{tabular}{lcccc}
\hline
Prior Modeling & SSIM$\uparrow$ & PSNR$\uparrow$ & LPIPS$\downarrow$ & NIQE$\downarrow$ \\
\hline
w/o Physical Prior &
0.79 &
22.90 &
0.21 &
4.22 \\
Isotropic Gaussians &
\textcolor{blue}{0.82} &
\textcolor{blue}{24.19} &
\textcolor{blue}{0.17} &
\textcolor{red}{3.94} \\
Anisotropic Gaussians (Ours) &
\textcolor{red}{0.84} &
\textcolor{red}{24.38} &
\textcolor{red}{0.12} &
\textcolor{blue}{4.14} \\
\hline
\end{tabular}
}
\end{table}

\begin{table}[t]
\centering
\caption{Ablation results of different training-objective configurations. Red and blue indicate the best and second-best results for each metric, respectively.}
\label{tab:training_objective_ablation}
\resizebox{\columnwidth}{!}{
\begin{tabular}{lcccc}
\hline
Objective Configurations & SSIM$\uparrow$ & PSNR$\uparrow$ & LPIPS$\downarrow$ & NIQE$\downarrow$ \\
\hline
w/o $\mathcal{L}_{col}$ &
0.81 &
23.67 &
0.21 &
4.28 \\
w/o $\mathcal{L}_{edge}$ &
0.80 &
22.81 &
0.27 &
4.35 \\
w/o $\mathcal{L}_{exp}$ &
0.82 &
23.01 &
0.20 &
4.19 \\
w/o $\mathcal{L}_{smooth}$ &
\textcolor{blue}{0.83} &
\textcolor{blue}{23.84} &
\textcolor{blue}{0.14} &
\textcolor{red}{4.07} \\
\textbf{Full Objectives} &
\textcolor{red}{0.84} &
\textcolor{red}{24.38} &
\textcolor{red}{0.12} &
\textcolor{blue}{4.14} \\
\hline
\end{tabular}
}
\end{table}

\subsection{Ablation Study}
\label{sec:ablation}

To validate the effectiveness of the key components and loss-function design in GLFS, we conduct systematic ablation studies on the LOL dataset. All model variants adopt the same training strategy and hyperparameter settings as the full model to ensure a fair comparison. Except for the ablation on training objectives presented at the end of Sec.~\ref{sec:ablation}, paired reconstruction supervision is retained in all other ablation settings. Depending on the ablation type, the corresponding architectural component or auxiliary regularization term is removed or replaced to evaluate its contribution beyond paired supervision.

\noindent\textbf{Validation of the Core Design.}
To investigate the individual contributions of the physics-based components, we perform fine-grained ablations on the Multi-Scale Gaussian Tokenizer (MGT), Gaussian Splatting Attention (GSA), and Gaussian Illumination Field (GIF). As shown in Table~\ref{tab:component_ablation}, the complete GLFS achieves the best overall performance. Removing MGT degrades the continuity of geometric details, while removing GSA results in the largest performance drop, indicating the importance of physics-guided long-range dependency modeling. Removing GIF disrupts the continuity of the gain field and degrades exposure uniformity.

Fig.~\ref{fig:component_ablation} further illustrates the effects of these components. The enhancement results in row (A) show that the complete model produces more natural and spatially consistent restoration. Row (B) visualizes the attention maps of the center token in Block~3. Compared with standard MHSA, which produces less structured attention patterns without explicit physical constraints, GSA introduces Gaussian bias to capture spatial structures consistent with illumination diffusion. The absolute error maps in row (C) further show that the complete GLFS produces lower reconstruction errors across the image.

\noindent\textbf{Effectiveness of the Core Architecture.}
To evaluate the effect of integrating 2DGS into the ViT architecture, we construct the following baselines for comparison: 1) \emph{Base-CNN}, which adopts the widely used ResNet-9blocks as the generator and serves as a convolutional baseline based primarily on local feature aggregation; 2) \emph{Base-ViT}, which replaces our Gaussian Splatting Attention with standard Multi-Head Self-Attention (MHSA) and excludes the Gaussian light field output head, representing a data-driven architecture with global modeling capability; and 3) \emph{GLFS (Ours)}, the complete physics-prior-driven ViT architecture proposed in this work.

As shown in Table~\ref{tab:architecture_ablation} and Fig.~\ref{fig:architecture_ablation}, Base-CNN produces more illumination artifacts and local overexposure under severe non-uniform illumination. It also obtains the lowest overall image-quality performance among the three architectures while having the largest parameter count (11.37\,M). Base-ViT improves the PSNR to 23.46\,dB by using global attention to model long-range dependencies. However, without explicit modeling of physical illumination degradation, residual noise remains in some extremely dark regions. Base-ViT also requires an inference time of 0.023\,s. GLFS further improves the PSNR to 24.38\,dB and reduces LPIPS to 0.12. As shown in Fig.~\ref{fig:architecture_ablation}, GLFS suppresses noise and illumination artifacts in dark regions while producing more spatially uniform restoration that is closer to the reference image. In addition, GLFS contains 7.80\,M parameters and requires an inference time of 0.014\,s, both lower than those of Base-ViT. These results indicate that combining Transformer-based non-local modeling with the continuous Gaussian light field prior improves restoration performance while maintaining moderate model complexity and inference cost.

\begin{figure*}[t]
\centering
\includegraphics[width=0.95\textwidth]{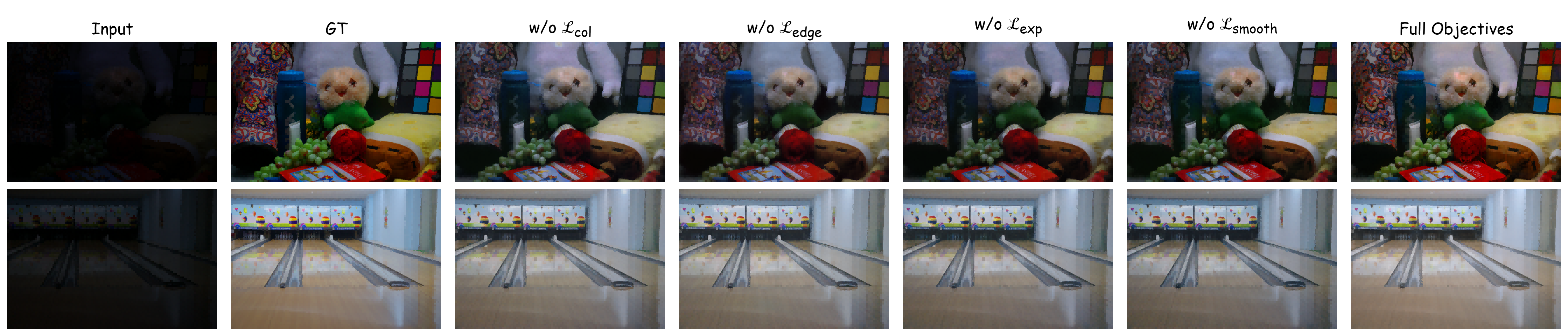}
\caption{Visual ablation comparisons of the training objectives. From left to right: low-light input, reference image, w/o $\mathcal{L}_{col}$, w/o $\mathcal{L}_{edge}$, w/o $\mathcal{L}_{exp}$, w/o $\mathcal{L}_{smooth}$, and Full Objectives.}
\label{fig:training_objective_ablation}
\end{figure*}

\begin{table}[t]
\centering
\caption{Ablation results of different Gaussian tokenization strategies. Red and blue indicate the best and second-best results for each metric, respectively.}
\label{tab:multiscale_tokenization_ablation}
\resizebox{\columnwidth}{!}{
\begin{tabular}{lccc}
\hline
Tokenization Strategy & SSIM$\uparrow$ & PSNR$\uparrow$ & Params (M)$\downarrow$ \\
\hline
Single-Scale ($\frac{1}{16}$) &
0.81 &
23.06 &
\textcolor{red}{6.43} \\
Single-Scale ($\frac{1}{4}$) &
\textcolor{blue}{0.82} &
\textcolor{blue}{23.81} &
\textcolor{blue}{7.02} \\
\textbf{Multi-Scale (Ours)} &
\textcolor{red}{0.84} &
\textcolor{red}{24.38} &
7.80 \\
\hline
\end{tabular}
}
\end{table}

\noindent\textbf{Impact of Physical Prior Modeling.}
The core of GLFS lies in inferring the spatial gain field $\Gamma(x)$ through the superposition of anisotropic Gaussian basis functions. To investigate how the formulation of this physical prior affects enhancement performance, we progressively simplify the illumination-field modeling mechanism: 1) \emph{w/o Physical Prior}, where Gaussian tokenization and the Gaussian Illumination Field (GIF) are removed and the network directly regresses an RGB gain map; 2) \emph{Isotropic Gaussians}, where each Gaussian covariance is constrained to an isotropic form, resulting in circular Gaussian kernels with adjustable scale and intensity; and 3) \emph{Anisotropic Gaussians (Ours)}, which uses the complete anisotropic Gaussian representation and allows elliptical Gaussian kernels with different orientations.

As shown in Table~\ref{tab:physical_prior_ablation}, removing the physical prior leads to a clear degradation in image naturalness, with NIQE increasing to 4.22. Without explicit light field modeling, the network directly predicts the gain map and is more prone to noise amplification in dark regions. Introducing isotropic Gaussians provides a spatial smoothness constraint, but their circular support cannot explicitly represent directional illumination patterns and shadow boundaries. By contrast, anisotropic Gaussians allow the covariance to adapt to different spatial orientations and achieve the best SSIM and LPIPS results in Table~\ref{tab:physical_prior_ablation}. These results indicate that direction-adaptive Gaussian basis functions provide a more suitable representation for spatially varying illumination fields.

\noindent\textbf{Ablation on Training Objectives.}
Among the nine main loss terms introduced in Section~\ref{sec:training_objectives}, the adversarial loss $\mathcal{L}_{adv}$ and cross-domain consistency loss $\mathcal{L}_{dc}$ constitute the basic framework for unpaired training, while $\mathcal{L}_{pair}$ provides absolute exposure and color references. We therefore focus on four constraints introduced to address key issues in low-light enhancement: 1) \emph{w/o $\mathcal{L}_{col}$}, where the color vector angular loss is removed and replaced with the conventional gray-world loss; 2) \emph{w/o $\mathcal{L}_{edge}$}, where the Y-channel Sobel-based edge loss is removed and replaced with a simple image-gradient loss; 3) \emph{w/o $\mathcal{L}_{exp}$}, where the unified exposure and spatial-consistency objective is removed; and 4) \emph{w/o $\mathcal{L}_{smooth}$}, where the TV smoothness objective for the physical field is removed. All four variants use the same paired supervision and hyperparameter settings as the full model.

\begin{figure}[t]
\centering
\includegraphics[width=\columnwidth]{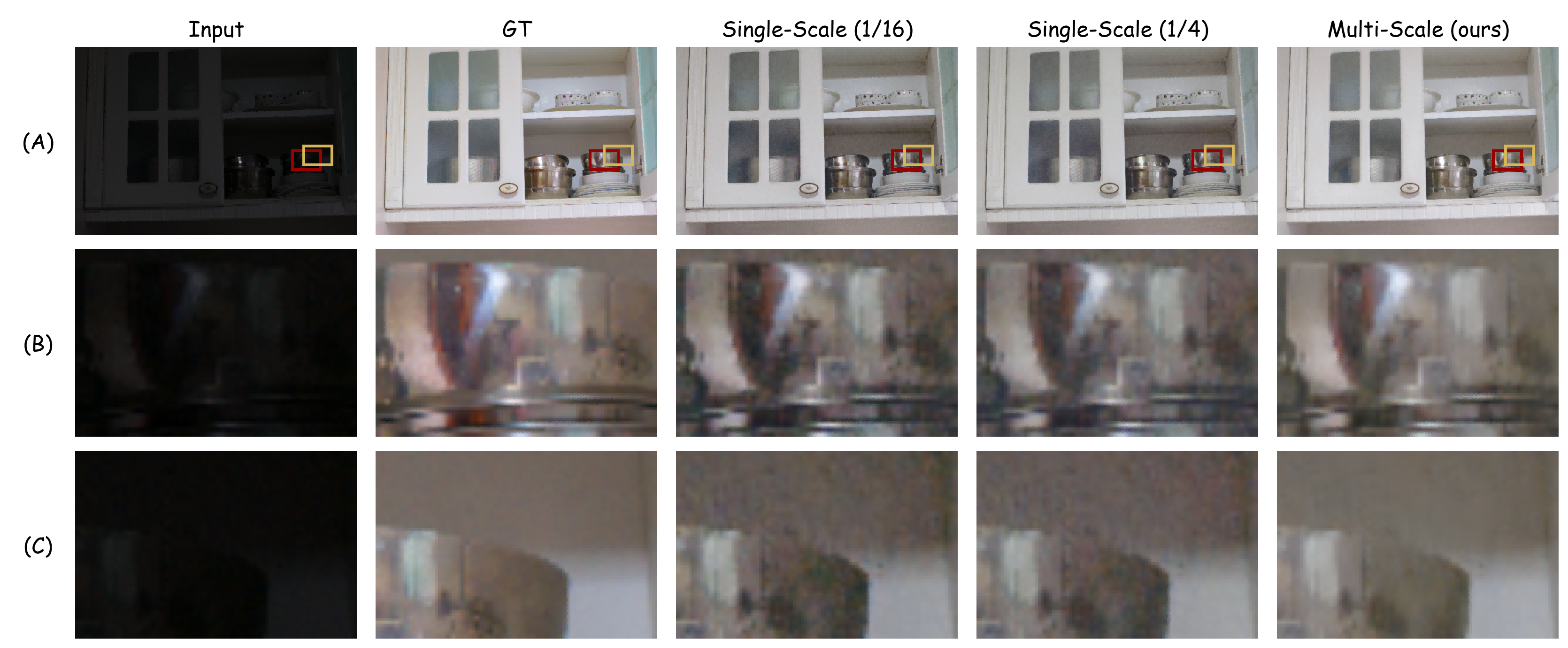}
\caption{Visual ablation comparisons of multi-scale Gaussian tokenization. (A) Full images. (B) Enlarged highlight regions. (C) Enlarged dark regions across illumination boundaries. The cropped regions are kept consistent across all three variants.}
\label{fig:multi_tokenization_ablation}
\end{figure}

As shown in Table~\ref{tab:training_objective_ablation}, the full objective achieves the best SSIM, PSNR, and LPIPS results. Removing $\mathcal{L}_{col}$ decreases PSNR to 23.67\,dB and increases LPIPS to 0.21. The conventional gray-world loss encourages the channel means toward a neutral balance, which can alter the color appearance of warm-toned scenes. In contrast, the angular loss constrains the direction of RGB color vectors without directly constraining their magnitudes, making it more compatible with brightness adjustment while preserving hue. Consistent with this observation, the first row of Fig.~\ref{fig:training_objective_ablation} shows an overall cooler color cast and reduced saturation for this variant.

Removing $\mathcal{L}_{edge}$ produces the largest performance degradation among the tested variants, with PSNR decreasing to 22.81\,dB, SSIM to 0.80, and LPIPS increasing to 0.27. This result suggests that applying the edge constraint to the Y channel helps preserve high-frequency structural information while reducing the influence of chromatic variations. The second row of Fig.~\ref{fig:training_objective_ablation} shows more noticeable blurring around the seam of the bowling lane when this loss is removed. When $\mathcal{L}_{exp}$ is removed, PSNR decreases to 23.01\,dB, accompanied by insufficient enhancement and spatially uneven brightness. These observations are consistent with the removal of the $16\times16$ local-mean exposure anchor and the four-directional spatial-consistency constraint.

Removing $\mathcal{L}_{smooth}$ yields the closest full-reference performance to that of the complete model among the four variants, although its SSIM, PSNR, and LPIPS remain inferior. This variant nevertheless achieves the lowest NIQE score of 4.07. The discrepancy between NIQE and the full-reference metrics indicates that these metrics respond differently to image statistics and reconstruction fidelity. Without the smoothness constraint, the network retains more high-frequency responses, including amplified noise in dark regions. Compared with this variant, the full objective increases NIQE by 0.07 while improving PSNR by 0.54\,dB and reducing LPIPS by 0.02, illustrating the trade-off introduced by the physical-field smoothness regularization.

\noindent\textbf{Effect of Multi-Scale Gaussian Tokenization.}
Illumination distributions generally contain both broad, diffuse global components and locally sharp highlights or deep shadows. To evaluate the contribution of multi-scale feature extraction and fusion in the proposed Multi-Scale Gaussian Tokenizer (MGT), we compare it with two single-scale strategies: 1) \emph{Single-Scale (Coarse only)}, which extracts Gaussian parameters only from low-resolution features at the $\frac{1}{16}$ scale; 2) \emph{Single-Scale (Fine only)}, which extracts Gaussian parameters only from high-resolution features at the $\frac{1}{4}$ scale; and 3) \emph{Multi-Scale (Ours)}, which jointly extracts and fuses Gaussian representations from both scales.

\begin{table}[t]
\centering
\caption{Trade-off analysis of model complexity, hyperparameter settings, and enhancement performance. Red and blue denote the best and second-best results for each metric, respectively.}
\label{tab:model_complexity}
\resizebox{\columnwidth}{!}{
\begin{tabular}{c c c c c c}
\hline
Blocks ($N_{\mathrm{blk}}$) & Heads ($H$) & Params (M) & FLOPs (G) & PSNR $\uparrow$ & SSIM $\uparrow$ \\
\hline
2 & 4 & 6.45 & 64.05 & \textcolor{blue}{24.24} & 0.82 \\
4 & 2 & 7.80 & 66.82 & 23.71 & \textcolor{blue}{0.83} \\
4 & 4 & 7.80 & 66.82 & \textcolor{red}{24.38} & \textcolor{red}{0.84} \\
4 & 8 & 7.80 & 66.82 & 23.18 & 0.82 \\
6 & 4 & 9.16 & 69.59 & 23.57 & 0.82 \\
\hline
\end{tabular}}
\end{table}

As shown in Table~\ref{tab:multiscale_tokenization_ablation}, both single-scale variants obtain lower full-reference performance than the complete model. When only the coarse scale is used, each Gaussian has relatively broad spatial support. This configuration captures the overall illumination distribution but has difficulty representing rapid variations around highlight and shadow boundaries. In contrast, using only the fine scale improves local fidelity but provides less global illumination context, leading to reduced consistency of the opacity field $\alpha$ over large homogeneous regions. The multi-scale design uses the coarse scale to model the overall geometry and global illumination distribution, while the fine scale complements it with high-frequency details, resulting in the best overall performance. Compared with the Coarse-only variant, Multi-Scale introduces 1.37\,M additional parameters while improving PSNR by 1.32\,dB, suggesting that the performance gain is associated with the complementary geometric information provided by the two scales rather than increased model capacity alone.

Fig.~\ref{fig:multi_tokenization_ablation} further illustrates these differences through local visual comparisons. In row (A), all three variants produce overall exposure close to that of the reference image, with the main differences concentrated in local regions. The enlarged highlight regions in row (B) show that Coarse-only produces more diffuse transitions and softened contours, consistent with its broader Gaussian support. Fine-only produces sharper boundaries but does not fully recover the intensity variations within the reflection. In contrast, Multi-Scale preserves both the highlight structure and a clear separation along the container boundary. In the enlarged dark regions around illumination boundaries in row (C), both single-scale variants exhibit color noise and block-like intensity fluctuations. These artifacts are accompanied by less continuous geometric guidance and localized gain amplification in dark regions, whereas Multi-Scale produces a smoother and more continuous luminance transition. Overall, these results indicate that multi-scale Gaussian tokenization improves both structural fidelity around highlight boundaries and the spatial continuity of enhancement gains in dark regions with only a modest increase in model complexity.

\noindent\textbf{Sensitivity Analysis of Model Complexity.}
Considering the constraints on parameter count and floating-point operations in practical deployment, the number of GS-ViT blocks ($N_{\mathrm{blk}}$) and attention heads ($H$) are two key architectural hyperparameters affecting model performance and computational characteristics. We evaluate $N_{\mathrm{blk}}\in\{2,4,6\}$ and $H\in\{2,4,8\}$, and report the number of parameters and FLOPs at an input resolution of $256\times256$ to identify an appropriate configuration.

As shown in Table~\ref{tab:model_complexity}, the performance exhibits a non-monotonic dependence on both hyperparameters. With $H=4$ fixed, increasing $N_{\mathrm{blk}}$ from 2 to 4 improves the PSNR from 24.24~dB to 24.38~dB. However, when $N_{\mathrm{blk}}$ is further increased to 6, the parameter count rises to 9.16~M while the PSNR decreases to 23.57~dB. This is because the magnitude of the residual update within each block is constrained, and excessively deep stacking may lead to accumulated drift in the Gaussian geometric parameters. With $N_{\mathrm{blk}}=4$ fixed, varying the number of attention heads does not change the parameter count or FLOPs in our implementation. When $H=2$, the limited number of independent temperatures $t_h$ restricts the parallel modeling of anisotropic structures. In contrast, when $H=8$, the per-head dimension $d_h$ becomes too small, weakening the representational capacity of the semantic dot-product term in Eq.~(\ref{eq:attention_logit}) and reducing the PSNR to 23.18~dB. Notably, the lightweight configuration with $N_{\mathrm{blk}}=2$ and $H=4$ already achieves 24.24~dB, indicating that the performance gain stems primarily from the explicit Gaussian light-field prior rather than from increased model capacity. Therefore, we adopt $N_{\mathrm{blk}}=4$ and $H=4$ in the final model.

\subsection{Limitations and Failure Case Analysis}

\begin{figure}[t]
\centering
\includegraphics[width=\columnwidth]{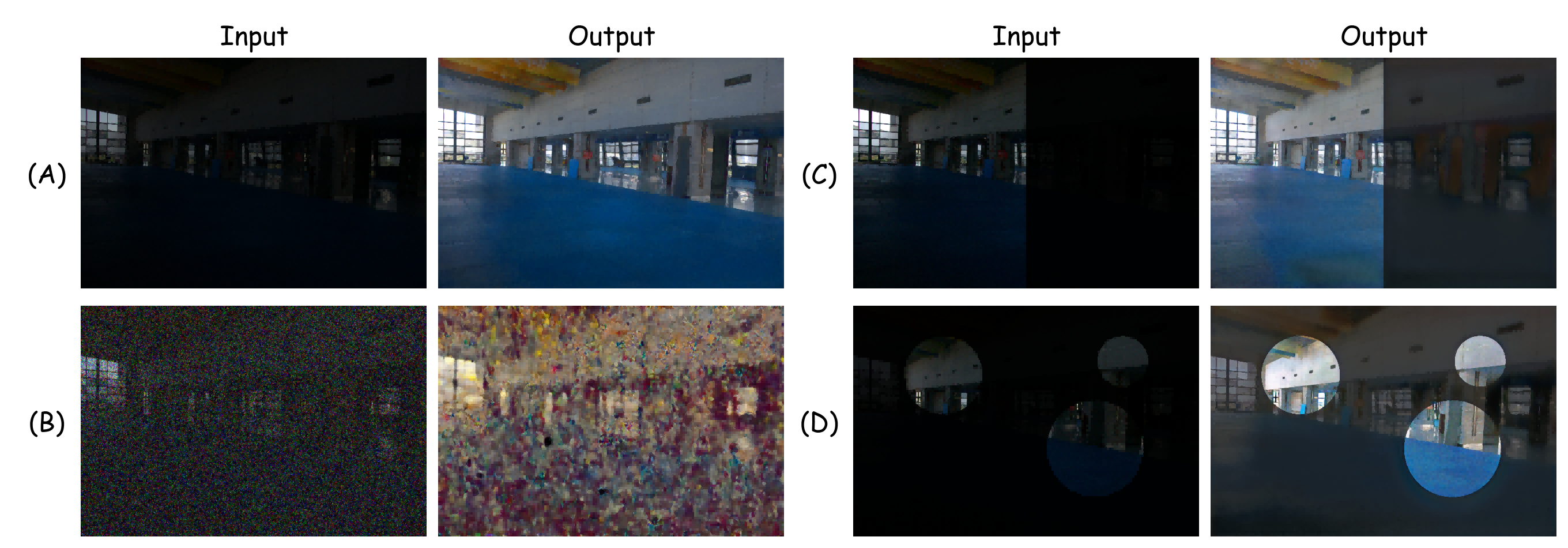}
\caption{Failure cases under extreme degradation conditions. (A) Inference on a conventional low-light image. (B) Under extreme sensor noise ($\sigma=100/255$), severe image degradation disrupts the semantic extraction process of the Gaussian tokenizer, resulting in structural artifacts. (C)--(D) Under abnormal illumination with hard boundaries, the inherent continuity and smoothness prior of the proposed Gaussian light field cannot perfectly fit extremely sharp illumination discontinuities, leading to over-smoothed boundary transitions or locally uneven exposure.}
\label{fig:failure_cases}
\end{figure}

Although the proposed GLFS model achieves strong performance on conventional low-light image enhancement tasks, its performance degrades under some extreme conditions. Table~\ref{tab:failure_cases} and Fig.~\ref{fig:failure_cases} present quantitative and visual analyses of representative failure cases involving severe noise and challenging illumination patterns.

\begin{table}[t]
\centering
\caption{Failure-case analysis under extreme degradation conditions. Red and blue indicate the best and second-best results for each metric, respectively.}
\label{tab:failure_cases}
\resizebox{\columnwidth}{!}{
\begin{tabular}{llcc}
\hline
Degradation Type & Sub-category & PSNR $\uparrow$ & SSIM $\uparrow$ \\
\hline
Baseline
& Clean
& \textcolor{red}{24.38}
& \textcolor{red}{0.84} \\

\hline
\multirow{3}{*}{Extreme Noise}
& $\sigma=25/255$
& \textcolor{blue}{16.26}
& 0.46 \\
& $\sigma=50/255$
& 14.12
& 0.31 \\
& $\sigma=100/255$
& 13.37
& 0.14 \\

\hline
\multirow{3}{*}{Abnormal Illum.}
& Circular (Severe)
& 12.37
& \textcolor{blue}{0.62} \\
& Multi-spot (Severe)
& 11.77
& 0.58 \\
& Stripes (Severe)
& 11.81
& 0.58 \\
\hline
\end{tabular}}
\end{table}

First, GLFS is sensitive to extremely strong sensor noise. When the noise level reaches $\sigma=100/255$, the PSNR drops to 13.37\,dB and the SSIM decreases to 0.14. As shown in Fig.~\ref{fig:failure_cases}(B), severe noise substantially corrupts the image content and interferes with Gaussian tokenization, making it difficult to extract reliable high-frequency geometric information. The resulting enhancement therefore contains noticeable structural artifacts and color blotches.

Second, GLFS exhibits limitations in modeling illumination with sharp spatial discontinuities. Under severe circular, multi-spot, and striped illumination patterns, the PSNR decreases to 11.77--12.37\,dB. As illustrated in Fig.~\ref{fig:failure_cases}(C)--(D), these cases are challenging for the continuous and smooth representation imposed by the Gaussian light field. When illumination changes abruptly across sharp artificial boundaries, the superposition of Gaussian basis functions may not represent such discontinuities sufficiently well, leading to over-smoothed transitions near illumination boundaries or uneven exposure in local regions. These observations motivate future work on adaptive, boundary-aware Gaussian light field modeling.

\section{Conclusion}

In this paper, we proposed GLFS, a physics-prior-driven Vision Transformer for low-light image enhancement under complex, non-uniform illumination. GLFS integrates the continuous light field representation of 2D Gaussian Splatting into the self-attention mechanism. By representing illumination as a superposition of anisotropic Gaussian basis functions, GLFS introduces an explicit physical bias for continuous modeling of the spatial gain field. Together with the hybrid supervision strategy, the color vector angular loss, and the luminance edge loss, GLFS improves exposure correction while preserving color consistency and structural details. Extensive experiments on six benchmark datasets demonstrate competitive overall performance across diverse degradation conditions.

Together with recent related studies~\cite{ref47,ref48,ref58}, these results suggest that Gaussian light field priors can provide structured physical guidance under different learning settings rather than being tied to a single training paradigm. Future work will further investigate robust Gaussian light field modeling under extreme sensor noise and adaptive boundary-aware representations for sharp illumination transitions, with the aim of improving robustness in more challenging real-world scenes.

\vspace{-20pt}
\begin{IEEEbiography}[{\includegraphics[width=1in,height=1.25in,clip,keepaspectratio]{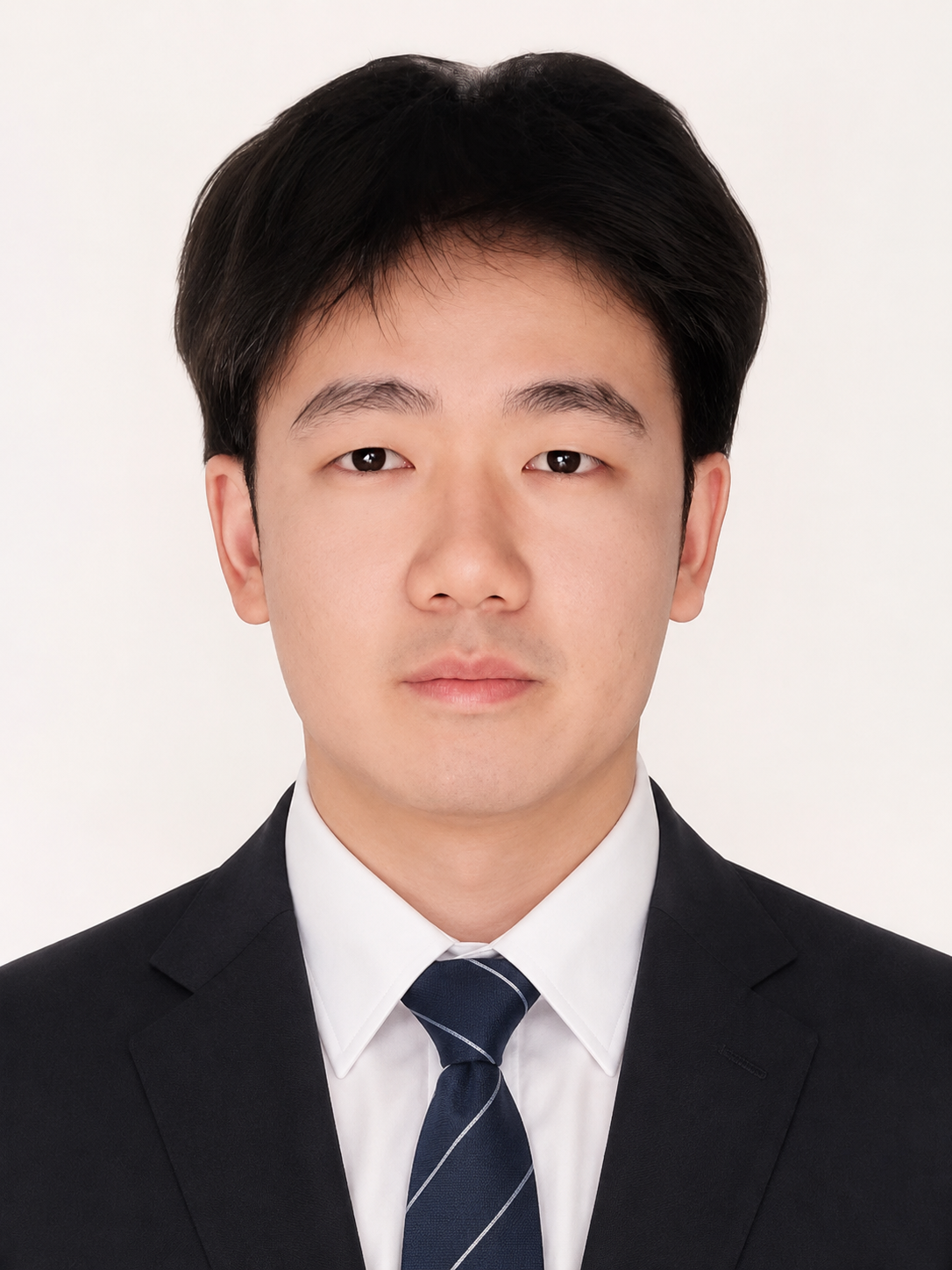}}]{Yuhan Chen}
received his master's degree in 2024 from the College of Mechanical Engineering at Chongqing University of Technology. He is currently pursuing the Ph.D. degree in College of Mechanical and Vehicle Engineering at Chongqing University, China. His research interests include deep learning, Low-level Vision and Gaussian Splatting.
\end{IEEEbiography}
\vspace{-40pt}

\begin{IEEEbiography}[{\includegraphics[width=1in,height=1.25in,clip,keepaspectratio]{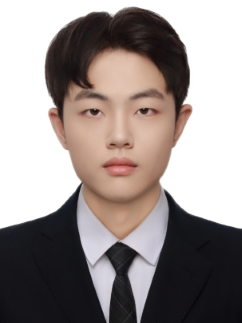}}]{Wenxuan Yu}
received the B.E. degree majoring in Mechanical Design, Manufacturing, and Automation at Chongqing University in 2025. He is currently pursuing the M.E. degree in Mechanical Engineering at Chongqing University, Chongqing, China. His research interests include computer vision, Gaussian Splatting and deep learning.
\end{IEEEbiography}
\vspace{-40pt}

\begin{IEEEbiography}[{\includegraphics[width=1in,height=1.25in,clip,keepaspectratio]{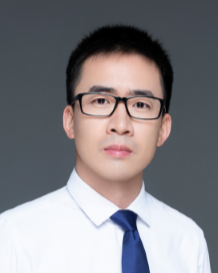}}]{Guofa Li}
received the Ph.D. degree in Mechanical Engineering from Tsinghua University, China, in 2016. He is currently a Professor with Chongqing University, China. His research interests include environment perception, driver behavior analysis, and smart decision-making based on artificial intelligence technologies in autonomous vehicles and intelligent transportation systems. He serves as the Associate Editor for IEEE Transactions on Intelligent Transportation Systems, IEEE Transactions on Affective Computing, and IEEE Sensors Journal.
\end{IEEEbiography}
\vspace{-20pt}

\begin{IEEEbiography}[{\includegraphics[width=1in,height=1.25in,clip,keepaspectratio]{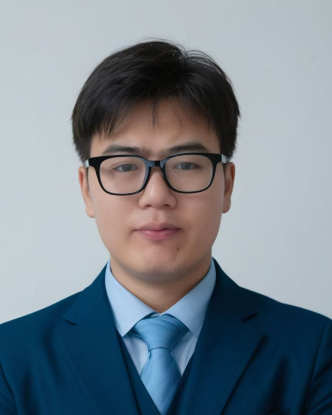}}]{Fuchen Li}
Fuchen Li is currently pursuing the M.S. degree in Electrical and Computer Engineering with the Herbert Wertheim College of Engineering, University of Florida, Gainesville, FL, USA. Prior to his graduate studies, he worked as an image algorithm engineer in the imaging industry, focusing on AI-ISP, HDR imaging, and auto white balance for embedded camera systems. His research interests include low-level vision, computational photography, vision-language models, and edge intelligence.
\end{IEEEbiography}

\vspace{-30pt}
\begin{IEEEbiography}[{\includegraphics[width=1in,height=1.25in,clip,keepaspectratio]{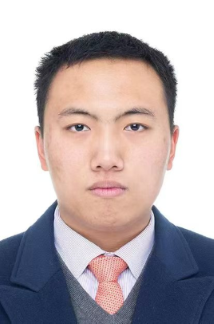}}]{Kunyang Huang}
is currently an masters degree student at Carnegie Mellon University. He previously interned at the Chongqing Institute of the Chinese Academy of Sciences and a research intern at the Changan Automobile Research Institute. His research interests include autonomous driving perception and deep learning methods.
\end{IEEEbiography}

\vspace{-15pt}
\begin{IEEEbiography}[{\includegraphics[width=1in,height=1.25in,clip,keepaspectratio]{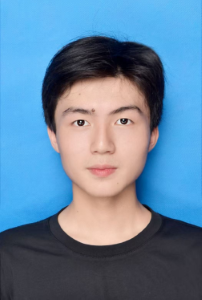}}]{Yicui Shi}
received the B.E. degree majoring in Automotive Engineering at Chongqing University in 2025. He is currently pursuing the M.E. degree in Automotive Engineering at Chongqing University, Chongqing, China. His research interests include computer vision, Gaussian Splatting and deep learning.
\end{IEEEbiography}

\vspace{-20pt}
\begin{IEEEbiography}[{\includegraphics[width=1in,height=1.25in,clip,keepaspectratio]{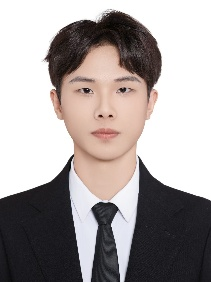}}]{Ying Fang}
received the B.E. degree majoring in Vehicle Engineering at Chongqing University of Technology. He is currently pursuing the M.E. degree in Mechanical Engineering at Chongqing University, Chongqing, China. His research interests include computer vision, Gaussian Splatting and deep learning.
\end{IEEEbiography}

\vspace{-20pt}
\begin{IEEEbiography}[{\includegraphics[width=1in,height=1.25in,clip,keepaspectratio]{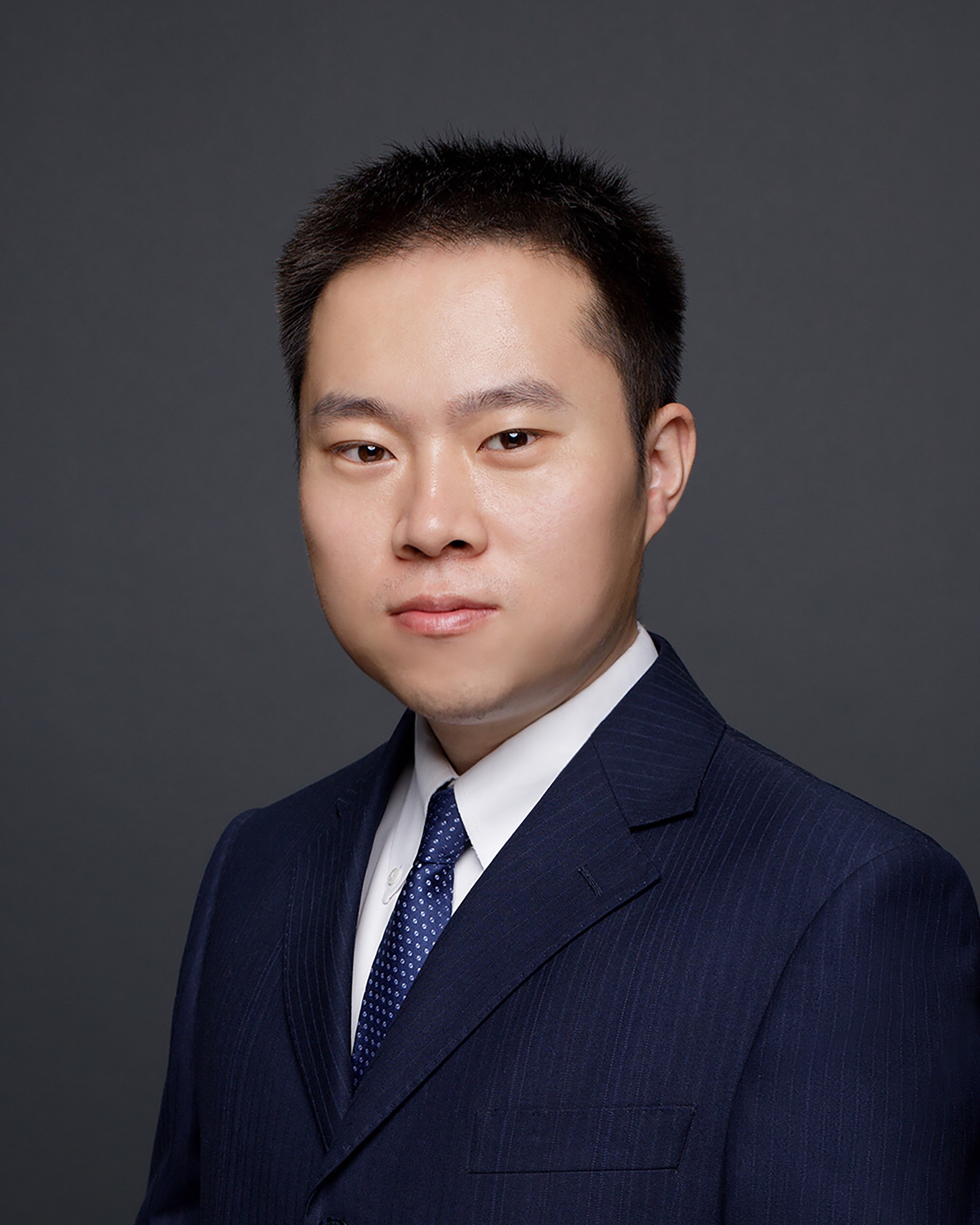}}]{Wenbo Chu}
received the B.S. degree in automotive engineering from Tsinghua University, Beijing, China, in 2008, the M.S. degree in automotive engineering from RWTH Aachen, Germany, and the Ph.D. degree in mechanical engineering from Tsinghua University, in 2014. He is with the Western China Science City Innovation Center of Intelligent and Connected Vehicles (Chongqing) Co., Ltd., Chongqing, China; the National Innovation Center of Intelligent and Connected Vehicles, Beijing, China; the Engineering Research Center of Mechanical Testing Tech. and Equip. Ministry of Education, Chongqing University of Technology, Chongqing, China.
\end{IEEEbiography}

\vspace{-20pt}
\begin{IEEEbiography}[{\includegraphics[width=1in,height=1.25in,clip,keepaspectratio]{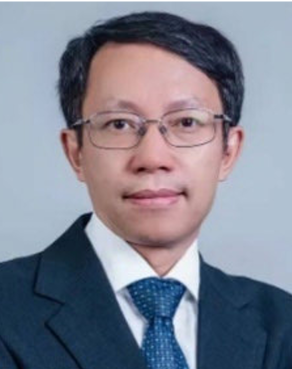}}]{Keqiang Li}
received the B.E. degree from Tsinghua University, Beijing, China, in 1985, and the M.E. and Ph.D. degrees from Chongqing University, Chongqing, China, in 1988 and 1995, respectively. He is currently a Professor with the School of Vehicle and Mobility, Tsinghua University. He is the Chief Scientist of Intelligent and Connected Vehicle Innovation Center of China, and the Director of State Key Laboratory of Automotive Safety and Energy of China. His current research interests include intelligent connected vehicles, cloud-based control for vehicles, and vehicle dynamics systems.
\end{IEEEbiography}

\vfill

\end{document}